\documentclass[11pt]{article}

\usepackage[margin=1in]{geometry}
\usepackage{booktabs}
\usepackage{makecell}
\usepackage{graphicx}
\usepackage{float}
\usepackage{hyperref}
\usepackage{xcolor}
\usepackage{CJKutf8}
\usepackage[most]{tcolorbox}
\usepackage{enumitem}
\usepackage{amsmath}
\usepackage{amssymb}

\hypersetup{
  colorlinks=true,
  linkcolor=blue,
  citecolor=blue,
  urlcolor=blue
}

\newcommand{\systemname}{MindMemOS}

\newcommand{\skillred}[1]{\textcolor{red}{#1}}
\newcommand{\skillblue}[1]{\textcolor{blue}{#1}}
\definecolor{iceblue}{RGB}{239,248,255}
\definecolor{midnight}{RGB}{25,44,71}
\definecolor{noahred}{RGB}{217,62,66}
\definecolor{noahredlight}{RGB}{255,244,244}
\definecolor{githubink}{RGB}{31,35,40}

\newcommand{\projectgithuburl}{https://github.com/mindscale-noah/MindMemOS}
\newcommand{\projectgithublink}{%
  \href{\projectgithuburl}{%
    \color{githubink}%
    \raisebox{-0.18ex}{\includegraphics[height=1.35ex]{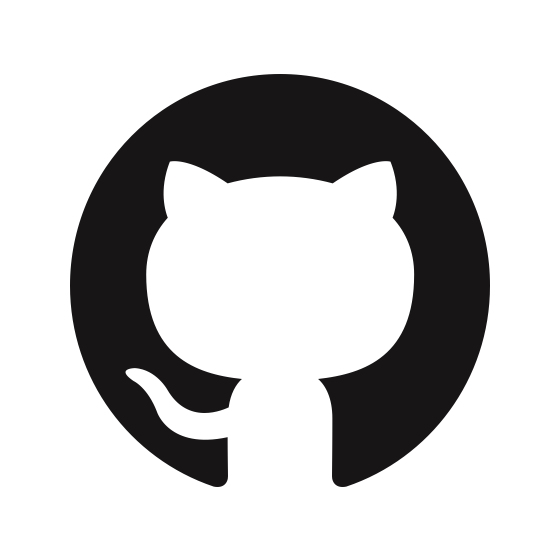}}%
    \hspace{0.35em}\nolinkurl{https://github.com/mindscale-noah/MindMemOS}%
  }%
}
\newcounter{promptcnt}
\newtcolorbox{promptbox}[3][]{%
    enhanced,
    colback=iceblue,
    colframe=midnight,
    fonttitle=\bfseries\sffamily\color{white},
    title={#2~\refstepcounter{promptcnt}},
    label={#3},
    attach boxed title to top left={xshift=5mm,yshift=-3mm},
    boxed title style={
        colback=midnight,
        sharp corners,
        boxrule=0pt,
        left=4mm,right=4mm,
        top=1mm,bottom=1mm
    },
    sharp corners,
    left=5mm,right=5mm,top=9mm,bottom=5mm,
    boxrule=0.6pt,
    shadow={2mm}{-2mm}{0mm}{black!20},
    breakable,
    before upper={\footnotesize\raggedright\setlength{\parindent}{0pt}\setlength{\parskip}{2pt}},
    #1
}

\newcommand{\tablesetup}{%
  \footnotesize
  \setlength{\tabcolsep}{4pt}%
  \renewcommand{\arraystretch}{1.15}%
}

\usepackage{algorithm}
\usepackage{algpseudocode}
\usepackage{caption}
\usepackage{tikz}
\usetikzlibrary{shapes, positioning, arrows.meta, fit, backgrounds, calc}

\title{MindMemOS: A Portable and Self-Evolving Memory Operating Layer for AI Agents}
\author{MindMemOS Project Team}
\date{\today}

\begin{document}

\thispagestyle{empty}
\vspace*{-1.0in}%
\hfill\includegraphics[height=1.0cm]{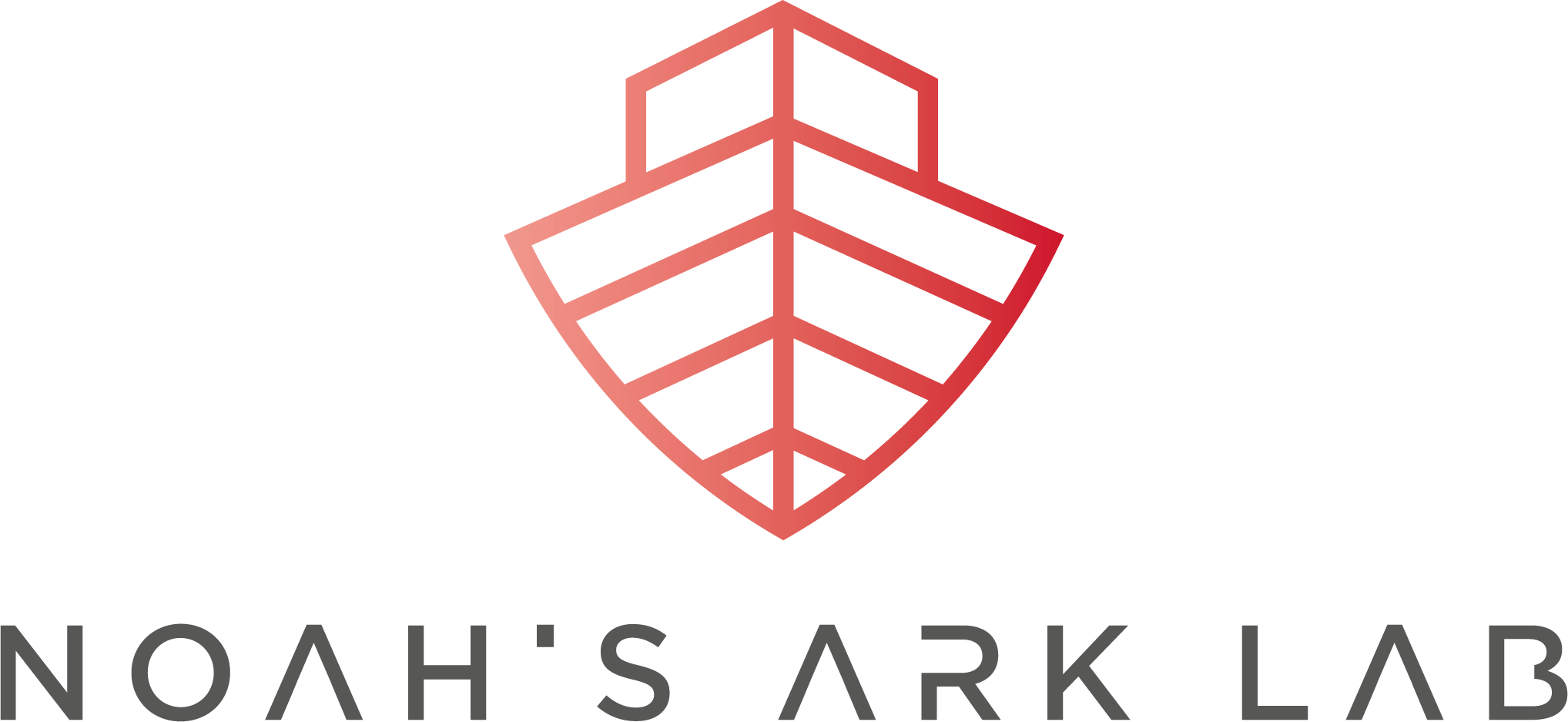}\hspace*{-0.9in}%
\vspace*{-0.5cm}

\begin{center}
    {\LARGE\bfseries MindMemOS: A Portable and Self-Evolving}\\[2pt]
    {\LARGE\bfseries Memory Operating Layer for AI Agents}\\[4pt]
    {\large MindMemOS Team}\\[1pt]
    {\normalsize Noah's Ark Lab, Huawei Technologies}
\end{center}

\vspace{-1pt}
{\color{midnight}\hrule height 1pt}
\vspace{2pt}

\begin{tcolorbox}[
    enhanced,
    colback=noahredlight,
    colframe=noahred,
    arc=3.5mm,
    outer arc=3.5mm,
    left=3mm, right=3mm, top=2.5mm, bottom=2mm,
    boxrule=0.7pt,
    before upper={\small\setlength{\parindent}{0pt}\setlength{\parskip}{0pt}},
]
{\centering\large\bfseries\sffamily\color{noahred}Abstract\par}
\smallskip

Memory is a core component of AI agents, enabling them to accumulate experience, maintain personalization, and adapt over long-term interactions. However, existing memory systems often remain fixed after development, limiting their ability to adapt their memory models, organization strategies, and procedural knowledge through continued use. We present \textbf{MindMemOS}, a portable and self-evolving memory operating layer that organizes open-world information using a unified entity–property–time structure. MindMemOS supports scenario-adaptive memory modeling, higher-order pattern discovery, autonomous memory refinement, and continuous skill evolution. Its MindMemEvolve algorithm employs validation-driven evolutionary search to optimize memory schemas for target scenarios, while dreaming consolidates accumulated memories by merging redundant records and resolving conflicts. In addition, implicit corrective feedback serves as a human-in-the-loop signal for identifying and revising potentially inaccurate or misaligned memories. Its MindSkillEvolve algorithm further transforms agent execution trajectories into reusable and progressively refined skills. MindMemOS achieves 94.03\% accuracy on LOCOMO and 70.63\% on PersonaMem. MindSkillEvolve improves SpreadsheetBench success by 9.2 percentage points over the initial-skill baseline.
\par\smallskip
{\centering\footnotesize\projectgithublink\par}
\end{tcolorbox}

\vspace{3pt}

\definecolor{baselineA}{RGB}{180,180,180}
\definecolor{baselineB}{RGB}{150,150,150}
\definecolor{baselineC}{RGB}{120,120,120}
\definecolor{baselineD}{RGB}{90,90,90}
\definecolor{mmvanilla}{RGB}{66,133,244}
\definecolor{mmschema}{RGB}{15,55,130}

\begin{figure}[H]
    \centering
    \includegraphics[width=\textwidth]{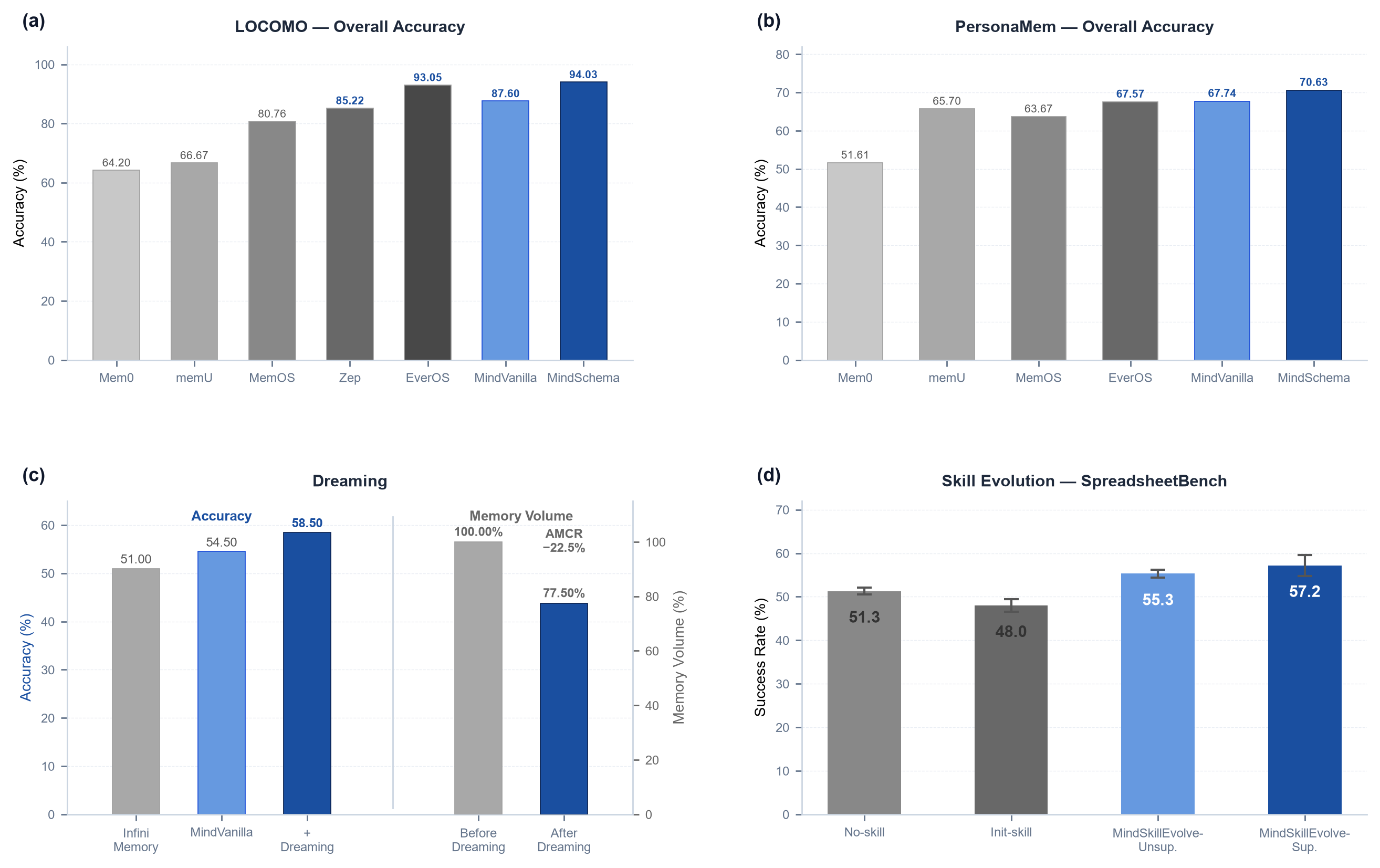}
    \caption{
        \textbf{(a)} LOCOMO --- Overall accuracy (\%). Gray: baselines. Blue: MindMemOS.
        \textbf{(b)} PersonaMem --- Overall accuracy (\%).
        \textbf{(c)} Dreaming (GPT-5-mini) --- Left: FactConsolidation overall accuracy. Right: active memory volume before vs.\ after dreaming (AMCR 22.5\%).
        \textbf{(d)} Skill Evolution --- SpreadsheetBench success rate (\%).
    }
    \label{fig:homepage-combined}
\end{figure}

\newpage


\section{Introduction}\label{introduction}


Large language model (LLM)-based agents are increasingly being integrated into software engineering, office productivity, information processing, and other knowledge-intensive workflows. Systems such as OpenClaw support task execution on personal computers~\cite{openclaw}, while coding-agent frameworks including Claude Code~\cite{claudecode}, OpenAI Codex CLI~\cite{codexcli}, and OpenCode~\cite{opencode} assist users with long-horizon software-development tasks. As these agents engage in repeated interactions, they must retain user preferences, previously acquired knowledge, task-relevant resources, and procedural experience beyond the capacity of a single context window. Memory therefore serves as an external information substrate that supports long-term personalization, knowledge reuse, and experience accumulation~\cite{rethinkingmemory,neurosciencememory,memorysurvey}.


Existing research on agent memory broadly follows two directions: memory models and memory systems. Memory models encode retained information within model parameters or latent representations. Parameter-based approaches incorporate customized information through adapters, fast parameters, or test-time training mechanisms~\cite{nestedlearning,fwpkm,memgen}, whereas latent-memory approaches maintain information in dedicated representation spaces, as exemplified by G-MemLLM~\cite{memorygate}, MemoryLLM~\cite{memoryllm}, G-Memory~\cite{gmemory}, and LatentMem~\cite{latentmem}. These approaches can improve memory-dependent reasoning, but they commonly rely on model-specific architectures or additional training procedures, which may limit their portability across foundation models and agent runtimes. Dynamic memory management approaches further optimize organization through meta-evolution, reinforcement learning, and skill-based mechanisms~\cite{memevolve,memrl,memskill}, but remain tightly coupled to specific training curricula or task environments, limiting cross-domain transferability.


Memory systems instead externalize information into textual or structured stores and use LLM-driven workflows for memory extraction, retrieval, and maintenance. Representative systems include Mem0~\cite{mem0}, memU~\cite{memu}, Zep~\cite{zep}, Mirix~\cite{mirix}, MemOS~\cite{memos}, EverOS~\cite{evermindos}, MemBrain~\cite{membrain} and VikingMem~\cite{vikingmem}. Some systems primarily emphasize the extraction and retrieval of factual memories, while experience-oriented approaches such as ExpeL~\cite{expel} and SCOPE~\cite{scope} focus on summarizing reusable task knowledge or optimizing procedural guidance. These systems have demonstrated the practical value of external memory, but their memory extraction policies, modeling structures, and organization strategies are often configured for particular scenarios and remain largely fixed after deployment ~\cite{unifyingmemory}.

This rigidity creates three challenges. First, an unstructured memory representation may be portable but provide limited support for fine-grained organization and temporal reasoning, whereas a manually designed schema can capture scenario-specific information but requires substantial adaptation when the target domain changes. Second, continuously accumulated memories may contain redundancies, outdated information, and conflicting statements, requiring systematic maintenance beyond online extraction and retrieval ~\cite{kang2026retainconsolidate,kang2026learningremember}. Third, memory content and procedural skills are often managed separately, making it difficult to transform accumulated execution experience into reusable and progressively refined agent capabilities.

To address these challenges, we present \textbf{MindMemOS}, a portable and self-evolving memory operating layer for AI agents. MindMemOS models open-world information using a unified entity–property–time structure that supports both modeling-free and schema-guided memory generation. A compact search module retrieves information through hybrid sparse–dense matching and bidirectional traversal over entity, property, relational, and temporal associations. On top of this representation, MindMemOS introduces four complementary evolution mechanisms. \textbf{MindMemEvolve} uses validation-driven evolutionary search to adapt memory schemas to target scenarios and discover task-relevant first-order and higher-order properties. \textbf{Dreaming} consolidates accumulated memories during offline periods by merging redundant records, resolving conflicts, and preserving provenance, while \textbf{Feedback} incorporates users’ corrective signals to identify and revise potentially inaccurate or misaligned memories. \textbf{MindSkillEvolve} further converts agent execution trajectories into reusable skill updates through unsupervised or score-guided refinement.

Our main contributions are as follows:

\begin{enumerate}

    \item \textbf{Scenario-Adaptive Memory Modeling.} We introduce an entity–property–time memory structure that organizes factual, relational, profile, and temporal information within a unified representation. The same system supports both open-domain vanilla ingestion and schema-guided extraction, allowing memory modeling to be adapted without changing the surrounding agent interface.
    

    \item \textbf{Active Memory Pattern Discovery.} We propose MindMemEvolve, an LLM-guided evolutionary search algorithm that optimizes memory schemas using task-specific evaluation signals. Through error-informed mutation, exploratory mutation, crossover, and selection, it adapts entity and property definitions and discovers higher-order patterns relevant to the target scenario.


    \item \textbf{Continuous Memory Refinement.} We introduce dreaming to consolidate accumulated memories during offline periods by merging redundant records, resolving conflicts, and preserving provenance. Complementarily, explicit and implicit feedback mechanisms use users’ corrective signals to identify and revise potentially inaccurate or misaligned memories while distinguishing transient task corrections from durable memory updates.


    \item \textbf{Experience-Driven Skill Evolution.} We introduce MindSkillEvolve, which continuously refines skills from accumulated usage experience. It analyzes agent execution trajectories to identify effective strategies and recurring failures, transforming them into versioned skill updates through unsupervised or score-guided evolution.

\end{enumerate}
\section{System Overview}

\subsection{System Architecture}

\begin{figure}[H]
    \centering
    \includegraphics[width=\textwidth]{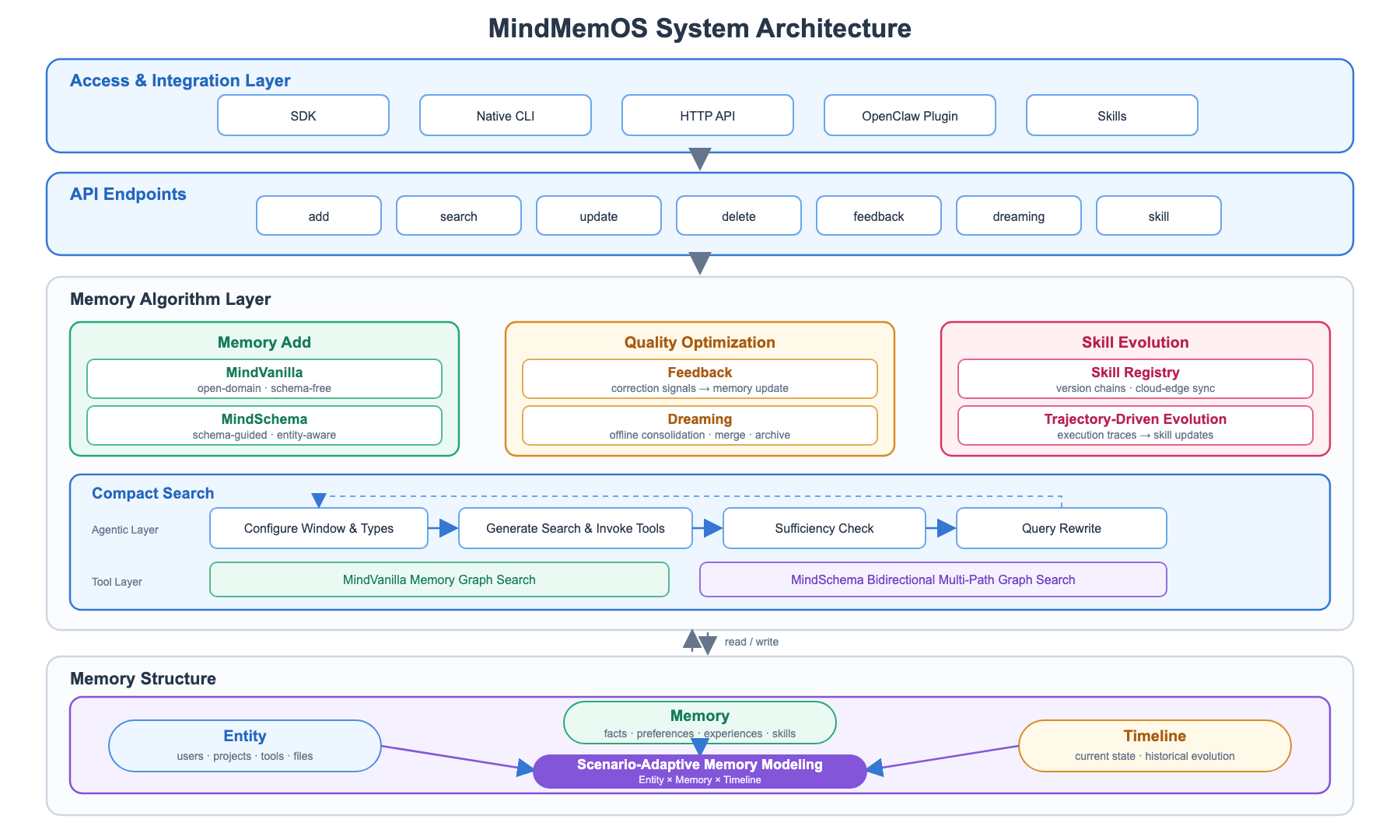}
    \caption{System architecture of MindMemOS.}
    \label{fig:system-arch}
\end{figure}

As shown in Figure~\ref{fig:system-arch}, MindMemOS adopts a layered architecture that decouples agent integration, memory algorithms, and memory structure (modeling and storage). At its foundation, the memory structure layer organizes memory modeling management, memory storage and skills. Built on this representation, the memory algorithm layer supports the complete memory lifecycle: MindVanilla policy and MindSchema policy, compact graph-based retrieval, feedback- and dreaming-driven quality optimization, and trajectory-driven skill evolution.

The agent and application layer expose these capabilities through a unified service abstraction, allowing MindMemOS to flexibly integrate with diverse agent frameworks via FastAPI-based HTTP APIs, SDKs, CLI commands, OpenClaw plugins or Skills while continuously optimizing memory content independently of the application runtime.

\subsection{Scenario-Adaptive Memory Modeling}


Memory modeling forms the foundation of the memory system, defining the data structures and associated algorithms for \texttt{add}, \texttt{search}, and related operations. As described in Section~\ref{introduction}, our design adopts scenario-adaptive memory modeling to organize open-world textual information into structured representations tailored to the target scenario.

Our memory model comprises three modeling dimensions:

\begin{figure}[H]
    \centering
    \includegraphics[width=\textwidth]{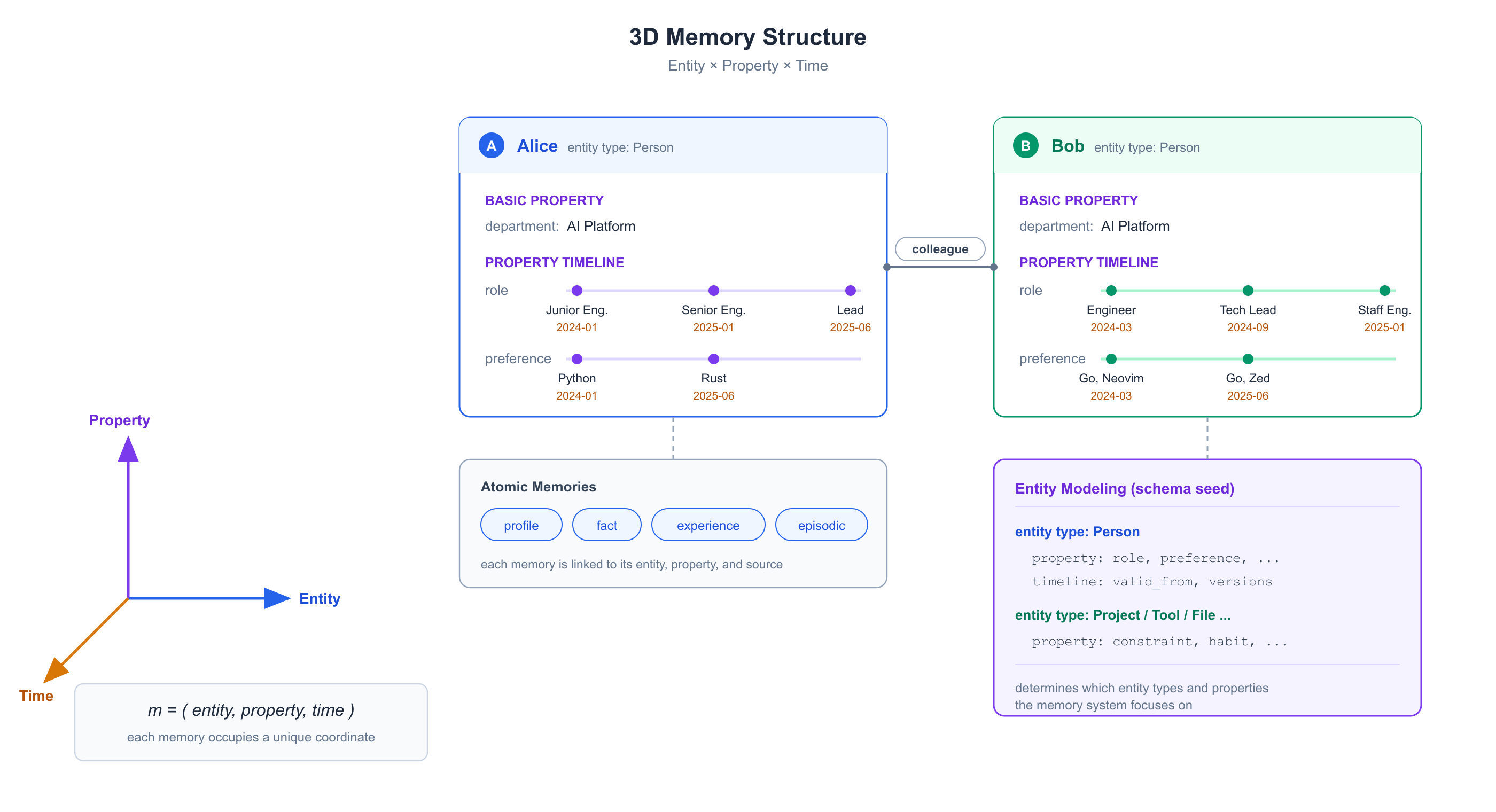}
    \caption{The 3D memory structure of MindMemOS, organized along the entity, property, and time dimensions.}
    \label{fig:memory-structure-3d}
\end{figure}

\paragraph{Three-dimensional Memory Structure.} Our memory system organizes memories in a three-dimensional (3D) graph defined by \textit{entity}, \textit{property}, and \textit{time}, thereby constructing and maintaining an information-grounded virtual representation of the real world. Each memory record is associated with an entity, a property describing a specific attribute or aspect of that entity, and a temporal reference indicating when the information holds or was observed. The graph further captures semantic relations among entities and temporal relations among successive records associated with the same entity--property pair. Together, these relational and temporal links support traceable navigation across interconnected entities and evolving information.

\paragraph{Entity Dimension.} The entity dimension identifies the subject described by a memory, such as a person, organization, location, event, or artifact. Entities serve as anchors for organizing information collected from different conversations, documents, and interactions. Relations between entities capture semantic connections such as participation, ownership, location, or social relationships.

\paragraph{Property Dimension.} The property dimension specifies the particular attribute or aspect of an entity described by a memory, such as a preference, activity, experience, relationship, or behavioral pattern. Under scenario-adaptive memory modeling, the set of properties is not restricted to a fixed universal schema; instead, property definitions can be adapted to the information needs of the target scenario. Properties may represent either explicit facts derived from individual observations or higher-order patterns synthesized across multiple observations.

\paragraph{Temporal Dimension.} The temporal dimension represents when information is valid, observed, or recorded. A temporal memory record binds an entity and a property to a specific content item and its temporal reference. Successive records associated with the same entity--property pair form a timeline, enabling the system to represent updates, changing preferences, and potentially conflicting versions without overwriting their historical context.

\section{Algorithm}

Building on the 3D memory structure, this section presents the core algorithms of MindMemOS, including two memory generation algorithms, compact retrieval, dreaming- and feedback-based memory refinement, offline memory self-evolution, and trajectory-driven skill evolution.

\subsection{Memory-Add: MindVanilla}




\paragraph{Turn-Aware Input Processing.}
MindVanilla generation processes dialogue and free-form text without relying on a predefined entity--property schema. It groups messages into turns based on conversational roles and temporal gaps, packs complete turns into token-bounded chunks, and compacts overlength turns when necessary. Each extractable message is normalized and linked to a source reference that preserves its role, timestamp, and original position.

\paragraph{Recall-Aware Memory Extraction.}
For each chunk, the system retrieves related active memories through content-hash matching, entity overlap when available, and BM25 retrieval, with candidates combined using weighted reciprocal rank fusion. The extractor receives a structured envelope that separates extractable evidence from contextual information: only the former may support new memory content, while history and recalled memories are used for disambiguation, duplicate detection, and conflict assessment. The resulting memories are represented as flat records and assigned coarse semantic types, such as profile, fact, episodic memory, tool trace, experience, or skill candidate.

\paragraph{Deduplication and Action Planning.}
Candidates produced across all chunks are deduplicated by content hash and memory type. After resolving their source references, a deterministic safety gate validates the candidates and maps them to \textsc{Add}, \textsc{Reinforce}, \textsc{Update}, \textsc{Merge}, or \textsc{Skip}. Accepted operations are vectorized as needed and stored together with their provenance and memory-relation links.


\subsection{Memory-Add: MindSchema}

We further elaborate on the dynamic generation process of the 3D memory structure based on entity modeling. As illustrated in Figure~\ref{fig:schema-guided-construction}, the memory generation pipeline comprises the following steps:

\begin{figure}[H]
    \centering
    \includegraphics[width=\textwidth]{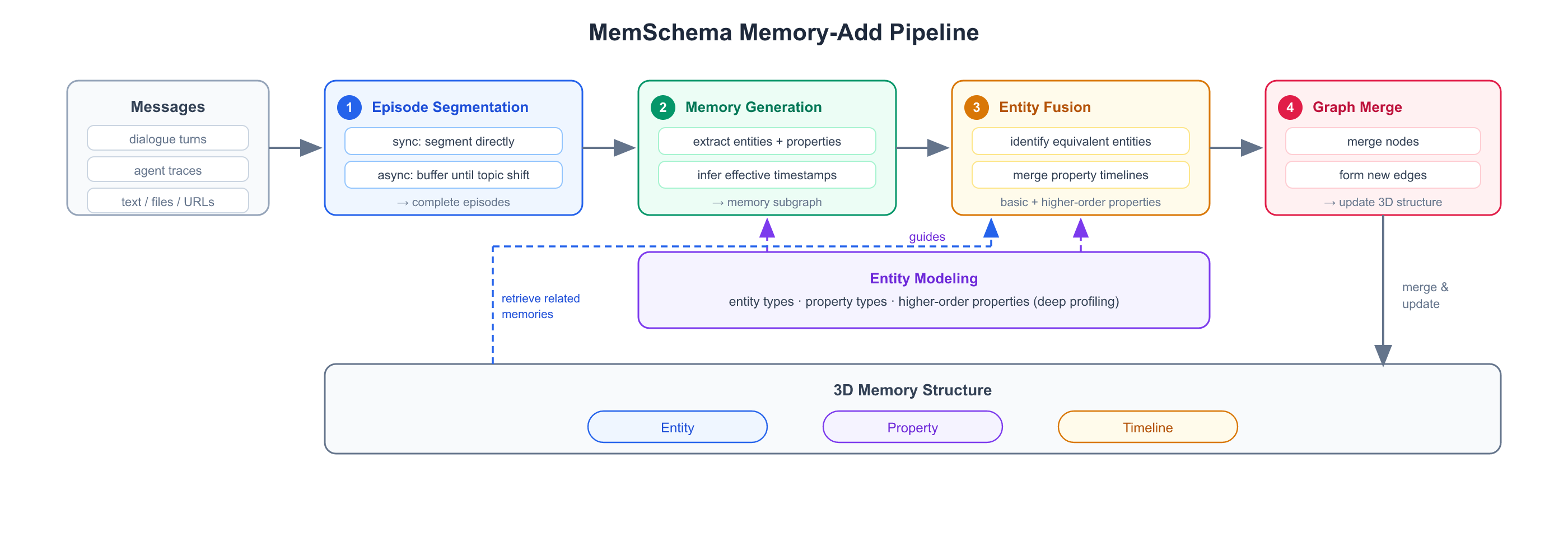}
    \caption{MindSchema memory generation pipeline: episode segmentation, memory generation, entity fusion, and graph merge, guided by memory modeling.}
    \label{fig:schema-guided-construction}
\end{figure}

\paragraph{Episode Segmentation.}
We employ an LLM-based memory segmenter to partition the message list received into complete sub-topic episodes. In synchronous mode, the entire message list from current memory-add requests is segmented directly. In asynchronous mode, messages first accumulate in a buffer until a topic shift is detected, forming a complete episode. It enables continuous accumulation and autonomous segmentation of memories, independent of message role transitions or session boundaries.

\paragraph{Memory Generation.}
When a memory-add request arrives, the system first maps the dialogue to the entity and property types defined in the scenario-adaptive memory schema. It then extracts the corresponding entity instances and property values from the source messages, resolving temporal expressions into absolute timestamps when sufficient temporal evidence is available. These outputs form the memory subgraph for the current episode. In addition, the system creates an episodic entity for each episode as a fallback representation for contextual information not captured by the schema-defined properties.

\paragraph{Entity Fusion.}
The system uses the current memory subgraph as a query to retrieve related memories and candidate matching entities from the memory store. It then resolves entity equivalence and merges each matched entity instance with its existing representation. The fusion process may update entity descriptions, insert or revise property values along their timelines, consolidate overlapping values, and mark outdated values as expired. If the scenario-adaptive schema defines higher-order properties for the entity type, the system additionally synthesizes these properties from relevant historical memories, such as the user's decision-making style or risk propensity.

\paragraph{Graph Merge.}
The system attempts to fuse the memory subgraph with the existing memory graph to form an updated memory graph. Through node merging (entity fusion above) and new entity edge relationship formation, the memory subgraph is ultimately merged and updated into the complete 3D memory structure, forming a structurally identical but information-enriched and refreshed memory graph.

\subsection{Compact Search}

\begin{figure}[H]
    \centering
    \includegraphics[width=\textwidth]{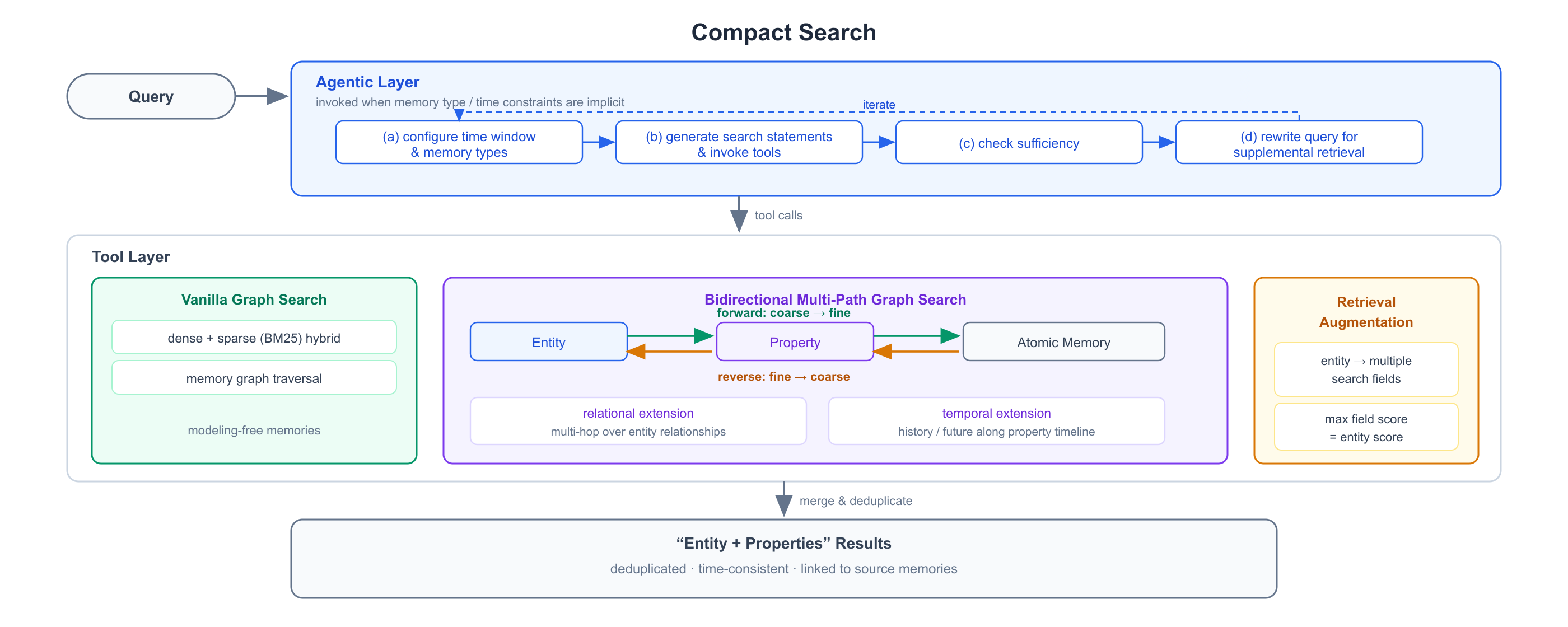}
    \caption{Compact search architecture: an outer agentic layer orchestrates an inner tool layer with bidirectional multi-path graph search over the 3D memory structure.}
    \label{fig:compact-search}
\end{figure}

We design a compact search module to improve retrieval coverage through multi-path traversal. Here, compact refers to restricting traversal to task-relevant entity, property, relational, and temporal associations rather than exhaustively expanding the memory graph. The overall search procedure is illustrated in Figure~\ref{fig:compact-search}.

\paragraph{Overall Architecture.}
The engine consists of an outer agentic layer for search planning and an inner tool layer for managing different search engines.

\paragraph{Agentic Search Planning.}
An LLM-based controller performs iterative retrieval planning. At each step, it determines the relevant temporal scope and memory types, formulates search queries, invokes retrieval tools, and assesses whether the collected evidence is sufficient. Based on this assessment, the controller either refines the retrieval plan or terminates the search.

\paragraph{Hybrid Bidirectional Graph Retrieval.}
The inner retrieval engine combines sparse lexical matching based on BM25 with dense semantic matching based on an embedding model, and fuses their ranked results using reciprocal rank fusion (RRF). This hybrid strategy is shared by the MindVanilla and MindSchema search tools. Over the 3D memory structure, the engine supports both forward traversal, which retrieves entities before selecting their relevant properties, and reverse traversal, which retrieves property-level memory records before tracing them to their associated entities. Results from both directions are deduplicated and organized into an entity-centered representation with the corresponding properties. Retrieved memories may be extended temporally through predecessor and successor property versions and relationally through associated entities. Relational expansion is optional and disabled in agentic mode to limit the retrieval scope during targeted multi-turn search. 

\paragraph{Retrieval Key Augmentation.}
To improve fine-grained entity matching, an entity description can also be decomposed into multiple search fields, each representing a distinct retrieval aspect. These fields are matched independently against the query, and the highest field-level score is used as the entity score, improving access to fine-grained evidence while preserving entity-level context.

\subsection{Dreaming}

\begin{figure}[H]
    \centering
    \vspace{-0.6em}
    \includegraphics[width=\linewidth]{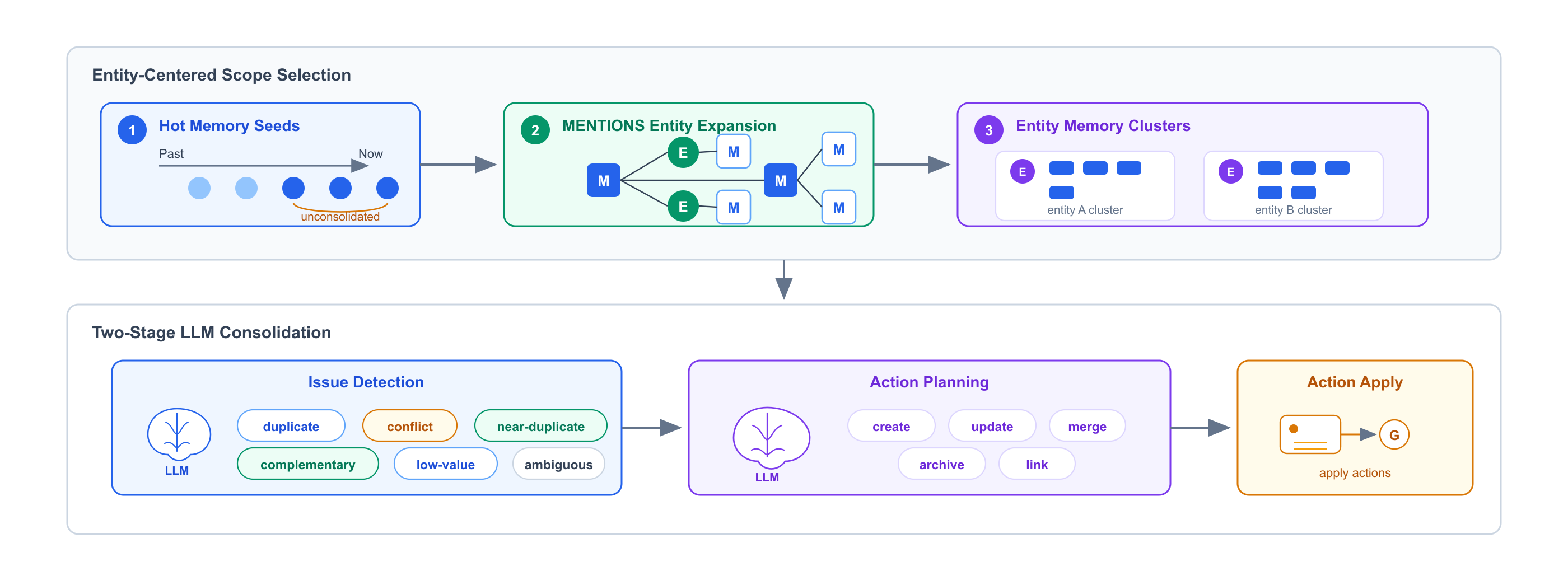}
    \vspace{-0.9em}
    \caption{Dreaming workflow in MindMemOS.}
    \label{fig:dreaming-flow}
    \vspace{-0.8em}
\end{figure}

Dreaming is the offline consolidation mechanism of MindMemOS. Online memory generation prioritizes the timely capture of useful facts and episodes, but incremental writes may leave redundant, overlapping, outdated, or conflicting records. Dreaming reorganizes these accumulated memories after interaction, reducing the maintenance burden on the online generation path.

\paragraph{Entity-Centered Scope Construction.}A dreaming run begins by selecting unconsolidated add records within a configurable lookback window. Each selected record serves as a seed for constructing a local consolidation scope: the system retrieves active memories associated with the same entity and groups them into an entity-centered cluster. This strategy confines consolidation to recently affected neighborhoods of the memory graph, avoiding both global comparison and isolated pairwise processing.

\paragraph{Issue Detection and Action Planning.}Each cluster is processed through a two-stage detect-then-act procedure. The first LLM call identifies focused issue groups involving conflicts, duplication, complementary fragments, low-value content, or ambiguous relationships. These groups are validated for subject consistency and sufficient evidence before further processing. The second LLM call converts each validated issue into a conservative mutation plan that may create a consolidated memory, update or merge existing records, archive obsolete information, or add explicit relational links. When entity identity, relational structure, or temporal order is uncertain, the planner favors non-destructive updates or links over archival.

\paragraph{Traceable Memory Consolidation.}The resulting mutation plan is applied through the standard memory write interface. Newly created or merged memories receive updated vectors and provenance links to their supporting records; where applicable, timeline edges connect successive versions within the same entity--property scope. After successful processing, the corresponding add records are marked as consolidated to prevent repeated handling in subsequent runs. Through this traceable process, dreaming can reduce redundancy, resolve conflicts, and preserve the lineage of consolidated information.


\subsection{Feedback}


Feedback provides an interaction-driven mechanism for converting corrective user signals into structured memory-maintenance actions. MindMemOS supports two complementary modes. \emph{Explicit feedback} is directed at the memory system and specifies what should be corrected, whereas \emph{implicit feedback} is inferred from corrections originally addressed to the task-performing agent. Both modes operate over relevant interaction context and candidate memories, but differ in how corrective evidence is obtained.

\begin{figure}[H]
    \centering
    \includegraphics[width=\linewidth]{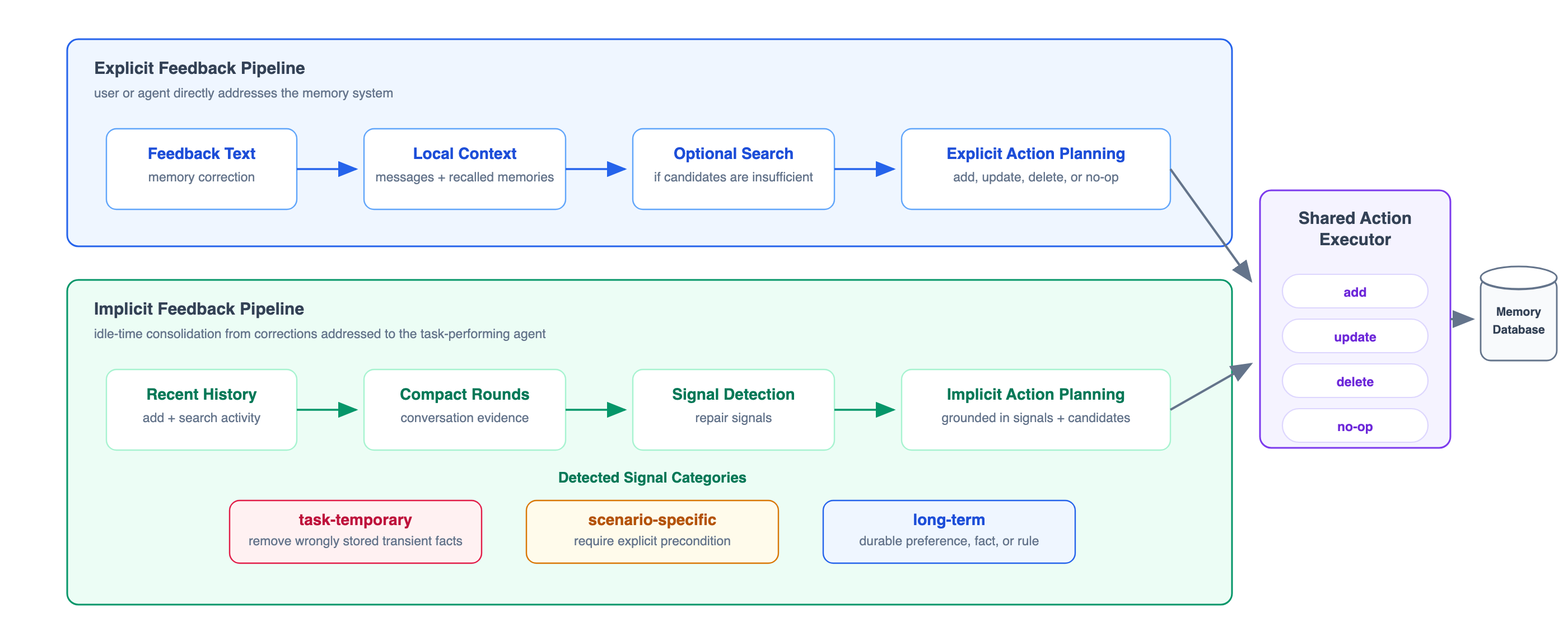}
    \caption{Feedback workflow in MindMemOS.}
    \label{fig:feedback-flow}
\end{figure}

\paragraph{Explicit Feedback.}

Explicit feedback allows users to revise stored memories through natural-language corrections rather than direct database manipulation. It takes the feedback statement, the relevant conversation context, and the initially retrieved memories as input. The conversation context helps disambiguate the correction target and distinguish durable information from instructions, while the retrieved memories provide candidate records for inspection and modification. When these candidates are insufficient to support a reliable decision, the planner may rewrite the feedback and its context into a search query, perform one supplemental MindVanilla search, and merge the retrieved records with the original candidates by memory ID. Based on the resulting evidence, the planner generates an \textsc{add}, \textsc{update}, \textsc{delete}, or \textsc{no-op} action and applies it through the shared memory-mutation interface. Corrections that are relevant only to the current task are excluded from durable storage, reducing the risk of propagating temporary instructions into long-term memory.

\paragraph{Implicit Feedback.}
In many interactions, users correct the task-performing agent rather than explicitly requesting a memory update. Such corrections may reveal inaccurate assumptions, revised preferences, rejected recommendations, or previously unstated constraints. Although these signals can be valuable for future interactions, they are easily lost after the current session because they are embedded in ordinary dialogue and may not directly reference any stored memory. Implicit feedback is designed to identify these corrective signals and selectively convert them into memory-maintenance actions, thereby reducing the recurrence of previously corrected behaviors.

The process runs asynchronously over recent session activity. It first compacts multi-turn interactions into semantically coherent rounds and detects correction signals from the user’s responses to the agent’s outputs. Each signal is then classified according to its persistence scope as \emph{task-temporary}, \emph{scenario-specific}, or \emph{long-term}. Task-temporary signals apply only to the current task and are not stored as new durable memories, although they may be used to correct or remove memories that have incorrectly retained one-off information. Scenario-specific signals are retained together with explicit applicability conditions, while long-term signals capture durable facts, preferences, or behavioral constraints that may generalize across future interactions. For each actionable signal, the system gathers relevant candidate memories and plans an \textsc{add}, \textsc{update}, \textsc{delete}, or \textsc{no-op} operation. This persistence-aware process reduces the risk of either discarding useful corrections or overgeneralizing transient feedback into unconditional long-term memory.



\subsection{MindMemEvolve: Validation-Driven Self-Evolution of Memory Schemas}

\begin{figure}[H]
    \centering
    \includegraphics[width=\textwidth]{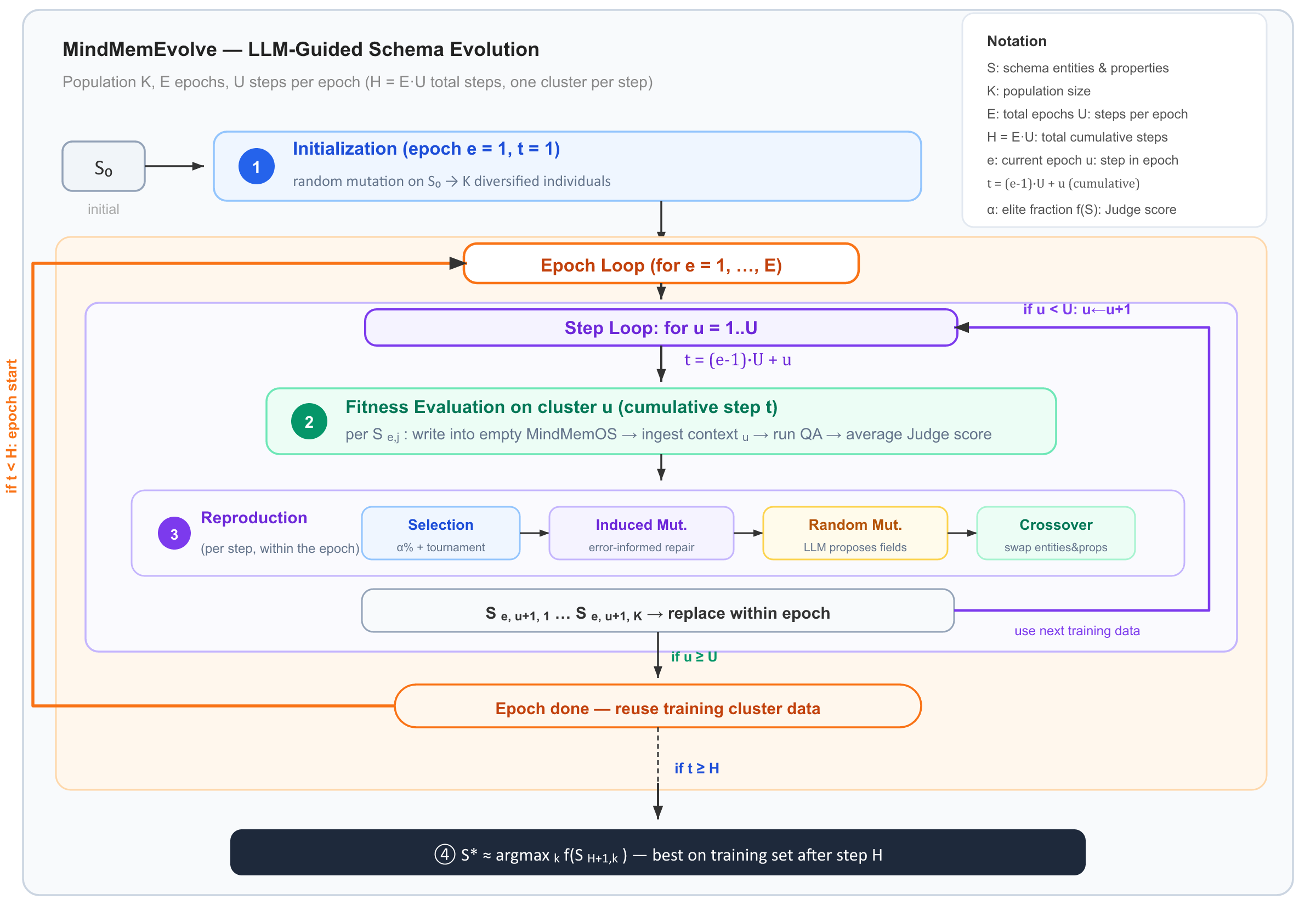}
    \caption{Overview of MindMemEvolve.}
    \label{fig:memevolve-flow}
\end{figure}

The core advantage of the memory modeling schema lies in its adaptability across scenarios, enabled by modular and configurable definitions of entities, first-order properties, and higher-order properties. This explicit representation makes the schema itself a natural target for optimization. To exploit this, we introduce \textbf{MindMemEvolve}, a self-evolution mechanism that automatically adapts an existing memory schema to a new target scenario.

In MindMemEvolve, the memory modeling schema is optimized with a training set $\mathcal{D} = \{(\mathbf{context}_i, q_i, a_i)\}_{i=1}^{N}$, where $\mathbf{context}_i$ denotes the context associated with the QA pair, $q_i$ a user query, and $a_i$ the reference answer. Note that different $q_i$ may share the same context, i.e., contexts can be the same across samples. Let $\mathcal{S}$ denote the candidate memory modeling schema being optimized. The optimization objective is to find the schema $\mathcal{S}$ that maximizes the Judge score:


\begin{equation}
\begin{aligned}
\mathcal{S}^*
= \arg\max_{\mathcal{S}}\;&
\frac{1}{N}\sum_{i=1}^{N}
\operatorname{Judge}\Bigl(
    a_i, \operatorname{LLM}\Bigl(
    q_i,\,
    \operatorname{MindMemOS}_{\mathrm{search}}
        (q_i;\mathcal{S})
    \,\Big|\, \\
&    \operatorname{MindMemOS}_{\mathrm{add}}
        (\mathbf{context}_i;\mathcal{S})
\Bigr)
\Bigr).
\end{aligned}
\label{eq:memevolve}
\end{equation}

where $\operatorname{Judge}(a,\hat{a}) \in [0,1]$ measures the correctness of the generated answer $\hat{a}$ with respect to the reference answer $a$. In each evaluation episode, $\text{MindMemOS}_{\text{add}}(\mathbf{context}_i; \mathcal{S})$ first processes $\mathbf{context}_i$ under schema $\mathcal{S}$ and writes structured memory entries (entities, properties, relationships) into the memory store. Given the query $q_i$, the search module $\text{MindMemOS}_{\text{search}}(q_i; \mathcal{S})$ retrieves a set of relevant memories. The retrieved memories are then provided to the LLM together with $q_i$ to generate the answer $\hat{a}_i$.

Since $\mathcal{S}$ is a discrete structured object, Eq.~(\ref{eq:memevolve}) defines a non-differentiable optimization problem. We adopt an LLM-guided evolutionary algorithm as a heuristic solver. The algorithm operates for $E$ epochs, each consisting of $U$ steps that iterate over distinct clusters of the training set, yielding $H = E \cdot U$ cumulative training steps. We index epochs by $e \in [1, E]$, steps within an epoch by $u \in [1, U]$, and cumulative steps by $t = (e-1) \cdot U + u$, with $t \in [1, H]$.

Let the population size be $K$. Denote the $j$-th individual at cumulative step $t$ as $\mathcal{S}_{t,j}$ ($t \in [1, H],\, j \in [1, K]$). The initial schema $\mathcal{S}_0$ is obtained either from a manually authored template or from an LLM-generated draft. The algorithm proceeds as follows:

\vspace{4pt}
\noindent\textbf{Step 1 --- Initialization.}
The first population $\{\mathcal{S}_{1,1}, \dots, \mathcal{S}_{1,K}\}$ is produced by applying random mutations to the initial schema $\mathcal{S}_0$, yielding $K$ diversified individuals.

\vspace{4pt}
\noindent\textbf{Step 2 --- Fitness Evaluation.}
At cumulative step $t$ (epoch $e$, cluster $u$), for each individual $\mathcal{S}_{t,j}$ we instantiate a sandboxed MindMemOS instance with a clean memory store, ingest the contexts of the $u$-th cluster under schema $\mathcal{S}_{t,j}$, and compute the fitness score $f(\mathcal{S}_{t,j})$ by evaluating Eq.~(\ref{eq:memevolve}) on the QA pairs belonging to that cluster.

\vspace{4pt}
\noindent\textbf{Step 3 --- Reproduction.}
The top $\alpha$ fraction of individuals (the elite) are preserved unchanged. The remaining $(1 - \alpha) K$ slots are filled by offspring produced through the following pipeline:

\begin{enumerate}
    \item \textbf{Parent selection.} $2(1 - \alpha) K$ parents are drawn via tournament selection (with possible duplicates).
    \item \textbf{Induced mutation.} The LLM analyzes error cases from the parent's evaluation, identifies information gaps, and proposes new entities or properties, or refines descriptions of existing ones.
    \item \textbf{Random mutation.} (a) The LLM proposes plausible entity types and property types for the target scenario, which are injected into the schema; (b)entities and properties whose frequency falls within the bottom $20\%$  are pruned with probability $p_{\text{prune}}$.
    \item \textbf{Crossover.} One parent serves as the base schema; entities and properties from the other parent that substantially differ from the base are inserted with probability $p_{\text{cross}}$.
\end{enumerate}

The resulting $K$ individuals form the population for step $t+1$, denoted $\{\mathcal{S}_{t+1,1}, \dots, \mathcal{S}_{t+1,K}\}$. Within an epoch, if $u < U$, the algorithm advances to the next cluster and returns to Step~2. When $u = U$, the epoch completes; if $t < H$, a new epoch begins at $u = 1$ and the clusters are iterated over anew.

\vspace{4pt}
\noindent\textbf{Step 4 --- Selection (Step $H+1$).}
After $H$ training steps, all $K$ individuals from the final population are evaluated on the training set. The individual with the highest fitness is selected as the approximately optimal schema:
\begin{equation}
\mathcal{S}^* \approx \arg\max_{j \in [1,K]} f(\mathcal{S}_{H+1,j}),
\label{eq:final_selection}
\end{equation}
where $\mathcal{S}_{H+1,j}$ denotes the $j$-th individual evaluated on the training set after step $H$.

The key insight of MindMemEvolve is that mutation operators are not purely random but are \emph{LLM-informed}: induced mutation leverages the Judge's error signal to perform targeted schema repair, while random mutation and crossover explore the schema space guided by the LLM's prior knowledge of the target domain.

\subsection{MindSkillEvolve: Trajectory-Driven Skill Evolution}

\begin{figure}[H]
    \centering
    \includegraphics[width=\linewidth]{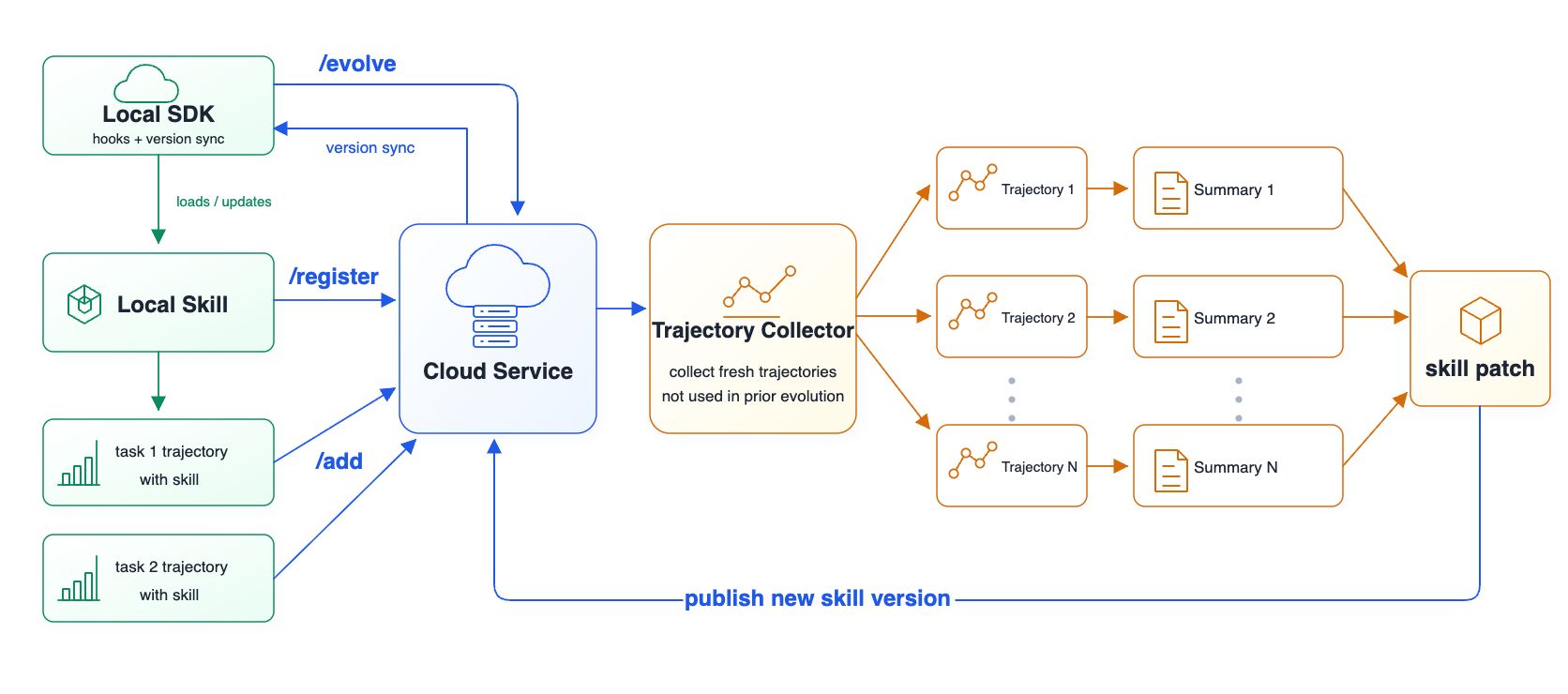}
    \caption{Skill registration and MindSkillEvolve workflow in MindMemOS.}
    \label{fig:skill-evolution-flow}
\end{figure}

\paragraph{Skill Registry and Lifecycle.}
MindMemOS provides cloud-edge collaborative lifecycle management of skills. The MindMemOS client is responsible for user-side skill registration, metadata collection, version synchronization, and rollback, while the cloud service provides centralized storage for skill content, maintains version chains, synchronizes skills across devices, and preserves historical version records. Through the SDK, a user can register a local skill by loading its content and collecting metadata such as its logical path, alias, content hash, and base version. The client then synchronizes the skill content and associated version metadata with the cloud service. For each registered skill, the cloud-side store maintains a version chain that supports historical version retrieval, rollback, and consistent skill states across devices, agents, and runtime environments.

 \paragraph{MindSkillEvolve.}
MindSkillEvolve, the skill-evolution algorithm in MindMemOS, is driven by real agent execution trajectories. MindMemOS reuses the Memory Add interface to collect task execution processes and provides SDK hooks through which different agent harnesses can attach skill context to each trajectory. The SDK hook layer is responsible for locating the skills loaded during the trajectory and matching them against registered skill versions, allowing MindMemOS to bind each trajectory to the specific skill versions it exercised. This design decouples MindSkillEvolve from how the underlying agent retrieves, loads, or executes skills, allowing it to adapt to different runtime patterns for skill usage. When MindSkillEvolve is triggered, the system collects unconsumed trajectories under the same skill version and aggregates them in chronological order. Each trajectory is first transformed into an evidence-driven analysis that preserves information relevant to skill improvement, including the task objective, execution process, key turning points, skill effectiveness, tool usage patterns, and final outcome.

Once the accumulated number of trajectories reaches a threshold, the system aggregates a batch of trajectory analyses, generates a skill-oriented improvement plan, and then uses an LLM to convert the plan into concrete edit operations on the SKILL. These edit operations are deterministically applied to the current skill content, producing a new cloud-side skill version.

MindSkillEvolve has two variants. \textbf{MindSkillEvolve-Unsup} relies solely on trajectory analyses to identify recurring successful strategies, failure patterns, and missing guidance. \textbf{MindSkill-\\Evolve-Sup} additionally uses trajectory scores as supervision signals to reinforce behaviors that consistently appear in high-scoring trajectories and suppress recurring errors found in low-scoring trajectories.

\section{Evaluation and Experiments}

\subsection{Dialogue-Centric Evaluation}

We evaluate MindMemOS on the two commonly used long-term-memory benchmarks in memory systems: \textbf{LOCOMO}, which assesses information extraction, effective recall, and simple derivative reasoning over multi-memory associations; and \textbf{PersonaMem}, which examines user profiling, preference analysis, and style-aware personalized recommendation. We compare two MindMemOS configurations---\textbf{MindVanilla}, a fast mode with dense/sparse retrieval, and \textbf{MindSchema}, a higher-accuracy schema-guided mode with scenario-adaptive memory modeling ---against strong, competitive baselines including Mem0~\cite{mem0}, memU~\cite{memu}, MemOS~\cite{memos}, and EverOS~\cite{evermindos}.

\begin{figure}[H]
    \centering
    \includegraphics[width=0.8\textwidth]{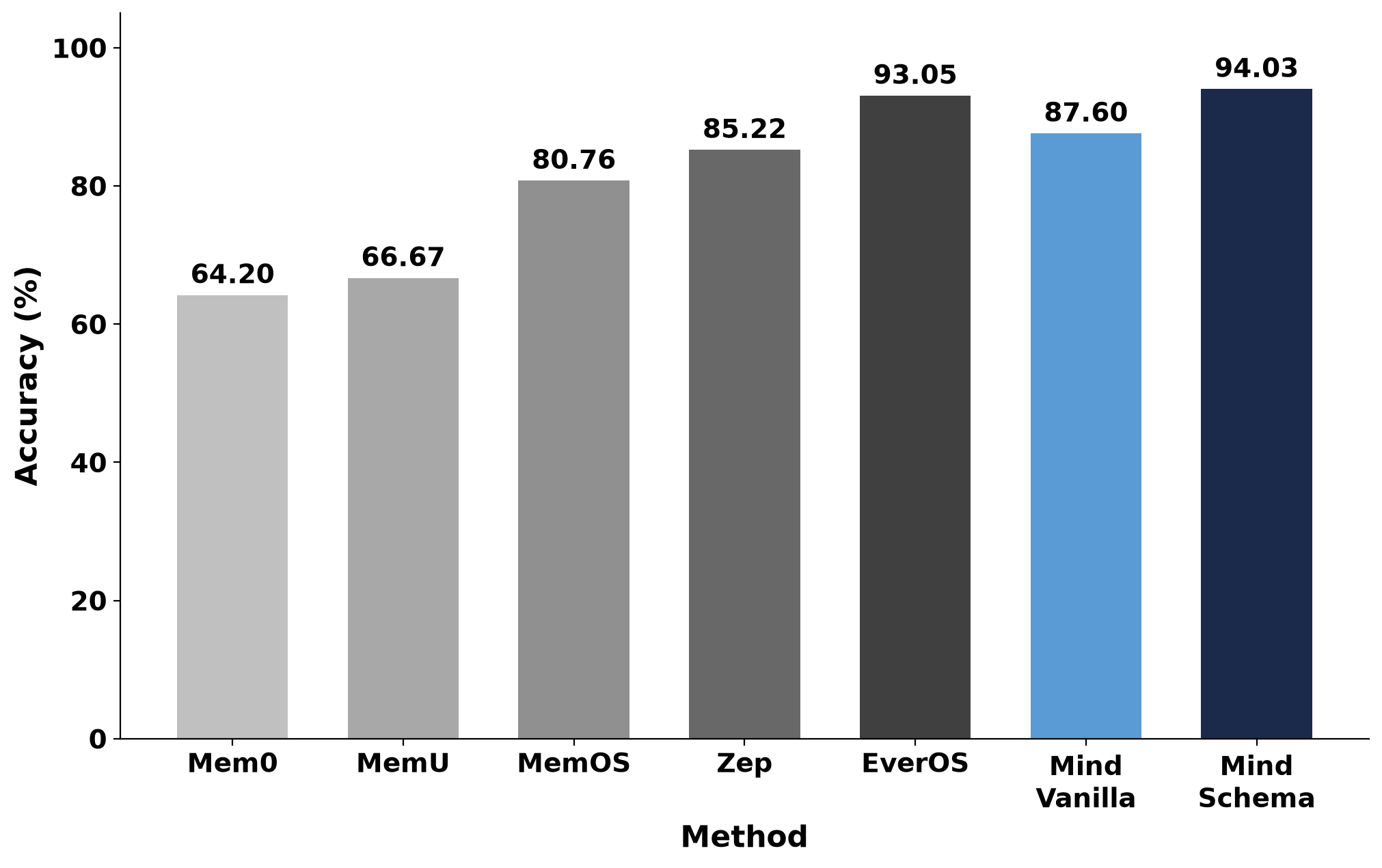}
    \caption{Overall accuracy comparison on LOCOMO. Baselines in gray, MindMemOS variants in blue.}
    \label{fig:locomo-overall}
\end{figure}

\paragraph{LOCOMO.}
LOCOMO~\cite{locomo} contains 10 long multi-session dialogues (averaging 300+ turns across up to 35 sessions) with 1,986 annotated questions across ten conversations, of which we focus on the four main reasoning types aligned with common research works~\cite{evermindos}. Following the evaluation protocol of EverOS~\cite{evermindos}, we report \textbf{per-category QA Accuracy (\%)} across four reasoning types---\textbf{Single-hop}, \textbf{Multi-hop}, \textbf{Temporal}, and \textbf{Open-domain}---along with the \textbf{Overall} accuracy. The answer model is gpt-4.1-mini. Baseline results for Mem0, memU, Zep, MemOS, and EverOS are cited from the EverOS paper; our experimental configuration, including the LLM driving the memory system, the embedding and reranking model, the answer model, and the judge model, is fully aligned with the EverOS default implementation. Results are shown in Figure~\ref{fig:locomo-overall} and Table~\ref{tab:locomo-eff}. MindSchema achieves the highest overall accuracy (94.03), ahead of EverOS (93.05) and Zep (85.22), with particular strength in Single-hop (96.79) and Multi-hop (93.97) reasoning. Notably, MindSchema achieves the highest Open-domain score (82.29) among all methods. MindVanilla (87.60) already outperforms MemOS (80.76) and Zep (85.22) by a clear margin, demonstrating that even the vanilla mode without modeling guidance provides competitive memory quality. Open-domain remains the most challenging category across all methods and is further influenced by the reasoning capability of the underlying answer model.

\begin{table}[htbp]
\centering
\tablesetup
\caption{LOCOMO --- Effectiveness. Per-category QA Accuracy (\%).}
\label{tab:locomo-eff}
\begin{tabular}{@{}lccccc@{}}
\toprule
\textbf{Method} & \textbf{Single-hop} & \textbf{Multi-hop} & \textbf{Temporal} & \textbf{Open-domain} & \textbf{Overall} \\
\midrule
Mem0            & 68.97 & 61.70 & 58.26 & 50.00 & 64.20 \\
memU            & 74.91 & 72.34 & 43.61 & 54.17 & 66.67 \\
MemOS           & 85.37 & 79.43 & 75.08 & 64.58 & 80.76 \\
Zep             & 90.84 & 81.91 & 77.26 & 75.00 & 85.22 \\
EverOS          & 96.67 & 91.84 & 89.72 & 76.04 & 93.05 \\
\midrule
\textbf{MindVanilla} & 92.03 & 85.82 & 83.80 & 66.67 & 87.60 \\
\textbf{MindSchema}  & 96.79 & 93.97 & 90.34 & 82.29 & 94.03 \\
\bottomrule
\end{tabular}
\end{table}

\paragraph{PersonaMem.}
PersonaMem~\cite{jiang2025know} is a long-range personalization benchmark comprising 20 different personas with alternating question answering. We retain the same experimental configuration as LOCOMO, except that the entity modeling (\texttt{entity\_modeling.json}) is designed around the user as the central entity and provides additional higher-order properties (see our code repository for details). Baseline methods are evaluated with local deployments of their open-source codebases, running under the same model configurations.
We report \textbf{per-category Accuracy (\%)} across seven query types---\textbf{Recall Sha.} (recalling user-shared facts), \textbf{Recall Mem.} (recalling facts mentioned by the user), \textbf{Track Evo.} (tracking full preference evolution), \textbf{Revisit} (revisiting reasons behind preference updates), \textbf{Suggest} (suggesting new ideas), \textbf{Recom.} (providing preference-aligned recommendations), and \textbf{General.} (generalizing to new scenarios)---along with the \textbf{Overall} accuracy. To ensure clarity regarding the evaluation metrics, we explicitly map the abbreviated query types used in our results to their corresponding definitions in the dataset. Specifically, \textbf{Revisit} corresponds to the task of \textit{recalling the reasons behind previous updates}.
MindSchema achieves the best overall accuracy (70.63), a gain of 3.06 percentage points over EverOS (67.57), driven primarily by improvements in Recall Sha. (81.40 vs.\ 74.42) and Suggest. (47.31 vs.\ 35.48). MindVanilla also achieves a high 67.74 overall, validating the effectiveness of the vanilla memory path even without explicit modeling guidance. Because the Recall Mem. category contains only 17 questions, a difference of one or two correct answers can produce a substantial change in category-level accuracy.

\begin{table}[htbp]
\centering
\tablesetup
\caption{PersonaMem --- Effectiveness. Per-category Accuracy (\%).}
\label{tab:personamem-eff}
\begin{tabular}{@{}lcccccccc@{}}
\toprule
\thead{Method} & \thead{Recall \\ Sha.} & \thead{Recall \\ Mem.} & \thead{Track \\ Evo.} & \thead{Revisit} & \thead{Suggest} & \thead{Recom.} & \thead{General.} & \thead{Overall} \\
\midrule
Mem0            & 46.51 & 41.18 & 65.47 & 90.91 & 12.90 & 34.55 & 43.86 & 51.61 \\
memU            & 64.34 & 64.71 & 66.20 & 87.88 & 31.18 & 67.27 & 84.21 & 65.70 \\
MemOS           & 53.49 & 82.35 & 66.91 & 79.80 & 41.94 & 69.09 & 75.44 & 63.67 \\
EverOS          & 74.42 & 64.71 & 64.03 & 85.86 & 35.48 & 65.45 & 84.21 & 67.57 \\
\midrule
\textbf{MindVanilla} & 76.74 & 88.24 & 65.47 & 87.88 & 17.20 & 80.00 & 82.46 & 67.74 \\
\textbf{MindSchema}  & 81.40 & 64.71 & 64.75 & 82.83 & 47.31 & 76.36 & 73.68 & 70.63 \\
\bottomrule
\end{tabular}
\end{table}

\begin{figure}[H]
    \centering
    \includegraphics[width=0.8\textwidth]{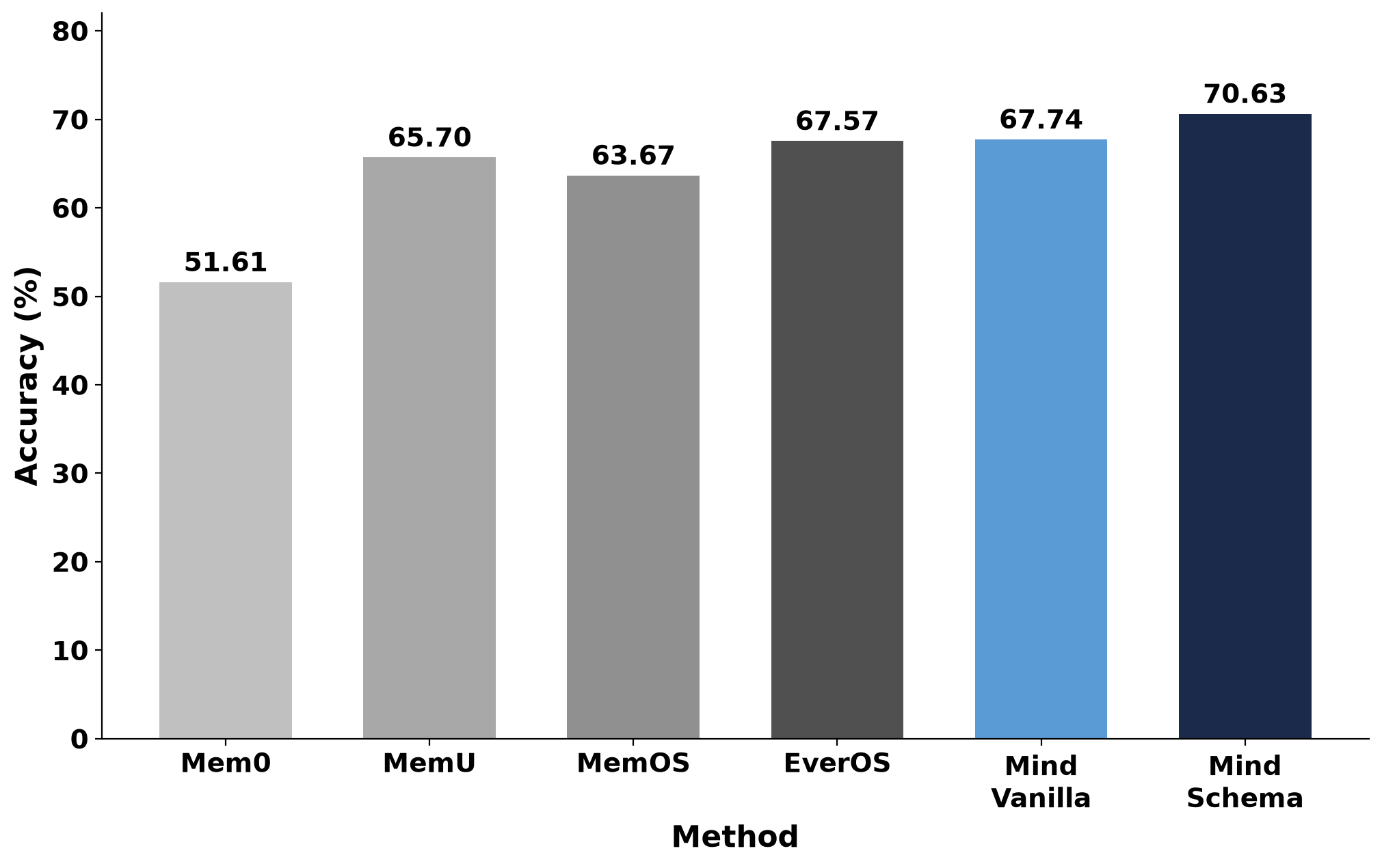}
    \caption{Overall accuracy comparison on PersonaMem. Baselines in gray, MindMemOS variants in blue.}
    \label{fig:personamem-overall}
\end{figure}

\subsection{Dreaming Evaluation}
We evaluated our dreaming algorithm on MemoryAgentBench~\cite{hu2026evaluating}. It is designed to evaluate memory agents, i.e., agent systems that incrementally store, update, and retrieve long-term information rather than consuming the entire context as a single static input. It organizes memory evaluation around four competencies: accurate retrieval, test-time learning, long-range understanding, and selective forgetting. This setting is well aligned with our evaluation because the agent receives information in temporal order and must maintain an effective memory state across updates. In particular, we use the FactConsolidation subset for selective forgetting, where an original fact is followed later by a contradictory rewritten fact. The benchmark treats the later fact as the valid memory and asks questions that require the agent to answer according to the final consolidated state. Thus, success on this subset requires not only retrieving relevant memories, but also resolving conflicts between outdated and newer information. We evaluate both single-hop (SH) and multi-hop (MH) variants across four context scales: 6k, 32k, 64k, and 262k.

\paragraph{Experimental Setup}
For each dataset split, we first construct memory with the MindVanilla algorithm and then evaluate question answering before and after running dreaming. The retrieval setting uses the fast search strategy with top-k $=50$. We report MindMemOS with gpt-4o-mini and gpt-5-mini using chunk size 4096, matching the settings used by the corresponding baseline memory systems. To isolate whether dreaming can resolve memory conflicts internally, timestamps are removed from the retrieved context during evaluation.

\paragraph{Experimental Results}

Following the official MemoryAgentBench evaluation protocol, Table~\ref{tab:dreaming-factconsolidation-results} reports the average Substring Exact Match (SubEM) accuracy across the four context scales. The active memory compression ratio (AMCR) denotes the fraction of active memories moved to the archive during dreaming. The Mem0~\cite{mem0}, MemoRAG~\cite{qian2025memorag}, and HippoRAG-v2~\cite{gutierrez2025ragmemorynonparametriccontinual} results are taken from MemoryAgentBench~\cite{hu2026evaluating} as reference, and we additionally compare against the hybrid retrieval variant of Infini Memory~\cite{ji2026infini} under gpt-5-mini. For gpt-4o-mini, dreaming improves the average single-hop accuracy from 0.635 to 0.738, the average multi-hop accuracy from 0.118 to 0.180, and the overall accuracy from 0.377 to 0.459, with AMCR values of 21.4\%, 19.4\%, and 20.4\%, respectively. For gpt-5-mini, dreaming further improves the average single-hop accuracy from 0.900 to 0.920, the average multi-hop accuracy from 0.190 to 0.250, and the overall accuracy from 0.545 to 0.585, with corresponding AMCR values of 23.5\%, 21.5\%, and 22.5\%. After dreaming, MindMemOS also outperforms Infini Memory on single-hop (0.920 vs.\ 0.800), multi-hop (0.250 vs.\ 0.220), and overall accuracy (0.585 vs.\ 0.510). These results show that dreaming consistently improves memory accuracy while compressing roughly one fifth of active memories, indicating that consolidation can reduce memory redundancy without degrading retrieval quality. A concrete example of this conflict resolution process is discussed in Section~\ref{dreaming_case}.

\begin{table}[t]
\centering
\caption{Average FactConsolidation accuracy and active memory compression ratio (AMCR) before and after dreaming.}
\label{tab:dreaming-factconsolidation-results}
\tablesetup
\begin{tabular}{@{}lcccccc@{}}
\toprule
& \multicolumn{2}{c}{\textbf{Single-Hop}} & \multicolumn{2}{c}{\textbf{Multi-Hop}} & \multicolumn{2}{c}{\textbf{Overall}} \\
\cmidrule(lr){2-3}\cmidrule(lr){4-5}\cmidrule(lr){6-7}
\textbf{Method}
& \textbf{Accuracy} & \textbf{AMCR}
& \textbf{Accuracy} & \textbf{AMCR}
& \textbf{Accuracy} & \textbf{AMCR} \\
\midrule
\multicolumn{7}{l}{\textit{gpt-4o-mini}} \\
Mem0 & 0.180 & -- & 0.020 & -- & 0.100 & -- \\
MemoRAG & 0.270 & -- & 0.070 & -- & 0.170 & -- \\
HippoRAG-v2 & 0.540 & -- & 0.050 & -- & 0.295 & -- \\
MindVanilla & 0.635 & -- & 0.118 & -- & 0.377 & -- \\
MindVanilla + Dreaming & 0.738 & 21.4\% & 0.180 & 19.4\% & 0.459 & 20.4\% \\
\midrule
\multicolumn{7}{l}{\textit{gpt-5-mini}} \\
Infini Memory & 0.800 & -- & 0.220 & -- & 0.510 & -- \\
MindVanilla & 0.900 & -- & 0.190 & -- & 0.545 & -- \\
MindVanilla + Dreaming & 0.920 & 23.5\% & 0.250 & 21.5\% & 0.585 & 22.5\% \\
\bottomrule
\end{tabular}
\end{table}

\subsection{Skill Evolution Evaluation}

\paragraph{Experimental Setup}

We evaluate MindSkillEvolve in a realistic client-cloud setting: the client executes real tasks in the user's local environment, while the cloud aggregates reported traces to improve the skill. In other words, MindMemOS performs delayed batch evolution and synchronizes the evolved skill back to the client for subsequent tasks.

We instantiate this evaluation on SpreadsheetBench-Verified\footnote{\url{https://huggingface.co/datasets/KAKA22/SpreadsheetBench/blob/main/spreadsheetbench_verified_400.tar.gz}}, a 400-task verified subset of SpreadsheetBench~\cite{spreadsheetbench}. Each task pairs a natural-language instruction with an initial workbook and a golden workbook for automatic verification. The subset contains 275 cell-level manipulation tasks and 125 sheet-level manipulation tasks, covering operations such as finding, extracting, summing, highlighting, removing, modifying, counting, deleting, calculating, and displaying spreadsheet content.

We compare four settings: \textbf{No-skill}, where the agent solves tasks without an external skill; \textbf{Init-skill}, where the agent uses the initial spreadsheet skill from SkillGrad~\cite{skillgrad}\footnote{\url{https://github.com/wwwhy725/SkillGrad/blob/main/seeds/xlsx/SKILL.md}} without evolution; \textbf{MindSkillEvolve-Unsup.}, where evolution uses only execution traces; and \textbf{MindSkillEvolve-Sup.}, where task scores are additionally used as supervision signals. Table~\ref{tab:sb_repeat_summary} reports the success rate, the consumption of agent tokens and the consumption of evolution tokens in three repeated runs. In our experiments, an evolution cycle is triggered every 40 tasks executed, with trajectories grouped into batches of eight tasks.

\paragraph{Experimental Results}

\begin{table}[t]
\centering
\caption{Results of MindSkillEvolve on SpreadsheetBench. Success rates are reported as the mean $\pm$ standard deviation over three runs.}
\label{tab:sb_repeat_summary}
\small
\setlength{\tabcolsep}{4pt}
\begin{tabular}{lcccc}
\toprule
Method & Success Rate  & Agent Tokens & Evolve Tokens \\
\midrule
No-skill & $51.3 \pm 0.8$ & $10.4\mathrm{M}$ &  - \\
Init-skill & $48.0 \pm 1.4$ & $16.9\mathrm{M}$ & - \\
MindSkillEvolve-Unsup. & $55.3 \pm 0.9$ & $27.3\mathrm{M}$ & $5.8\mathrm{M}$ \\
MindSkillEvolve-Sup. & $57.2 \pm 2.4$ & $25.2\mathrm{M}$ & $5.5\mathrm{M}$ \\
\bottomrule
\end{tabular}
\end{table}

The results show that skill evolution consistently improves task success rate over both no-skill execution and the unevolved initial skill. MindSkillEvolve-Unsup. improves success rate from $51.3\%$ to $55.3\%$, indicating that execution traces alone already provide useful signals for refining procedural spreadsheet knowledge. MindSkillEvolve-Sup. further improves success rate to $57.2\%$ by incorporating task scores as supervision. Notably, Init-skill performs worse than No-skill, which is consistent with the observation in SkillGrad~\cite{skillgrad}: an unoptimized initial skill may introduce misleading procedures or unnecessary constraints before it is adapted to the target task distribution.

\section{Case Studies}

\subsection{Dreaming}
\label{dreaming_case}




We examine a representative case from the MemoryAgentBench~\cite{hu2026evaluating} conflict-resolution benchmark. This benchmark evaluates whether a memory system can update its effective belief state when a newer memory contradicts an earlier one. The benchmark defines the most recently ingested fact as the ground truth, even when that fact conflicts with real-world knowledge. For the query \textit{What is Nobuhiro Watsuki famous for?}, the dataset expects \textit{The Fairly OddParents} under its temporal ordering. Before dreaming, the answer model instead returned \textit{Rurouni Kenshin}; after dreaming, it returned the expected answer. The case therefore tests temporal memory consolidation rather than factual recall from the base language model.

\begin{figure}[H]
    \centering
    \includegraphics[width=\linewidth]{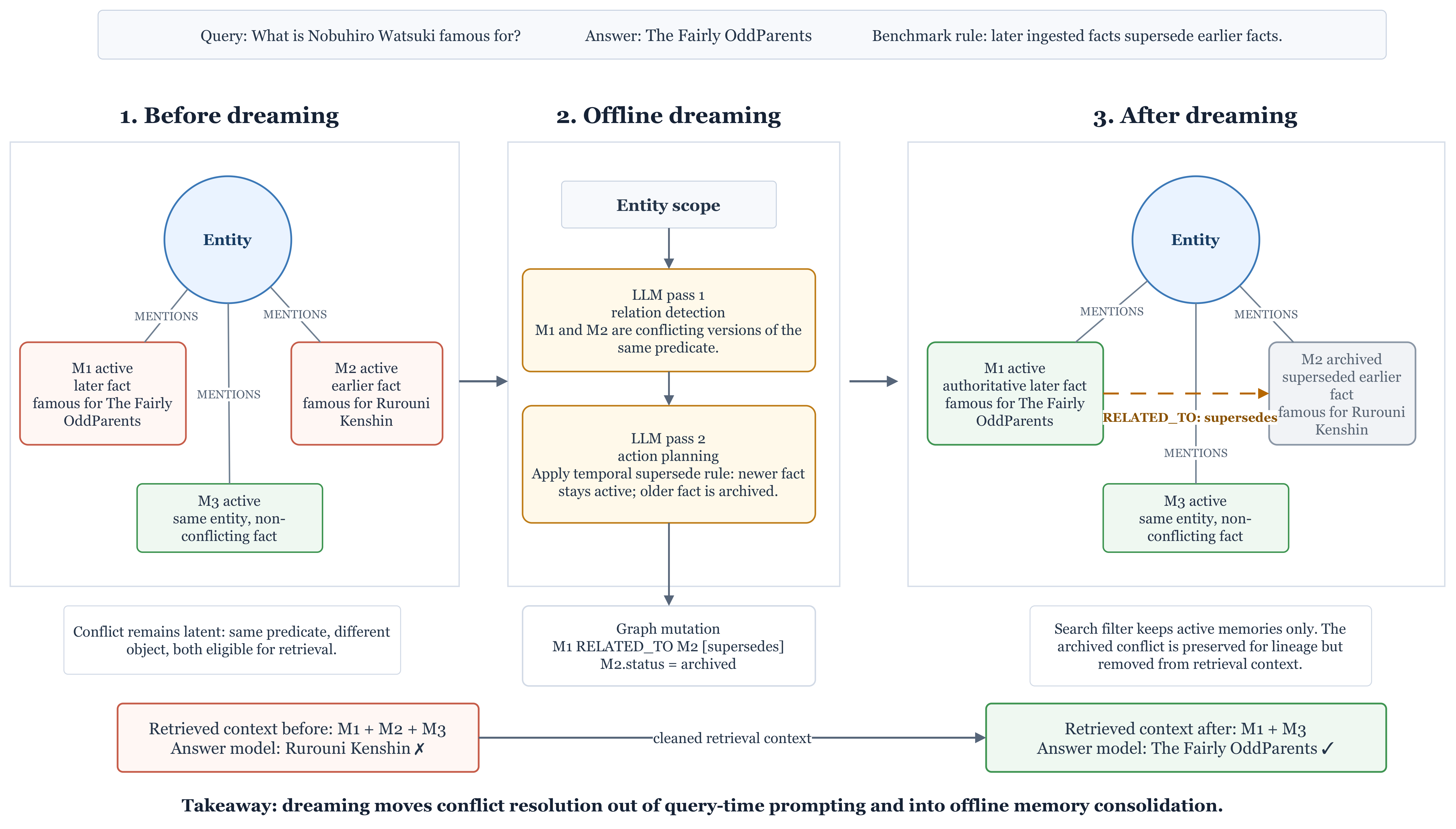}
    \caption{Conceptual overview of the dreaming pipeline. Before dreaming, conflicting active memories can be retrieved together. Dreaming detects the conflict within an entity-centered scope, archives the superseded memory, and records a \texttt{supersedes} relation so that subsequent retrieval returns an unambiguous active context.}
    \label{fig:dreaming-case-overview}
\end{figure}

\paragraph{Memory state before dreaming.}

The left panel of Figure~\ref{fig:dreaming-case-overview} shows the accumulated memory state before dreaming. The entity \texttt{nobuhiro\_watsuki} is connected to three active memories through \texttt{MENTIONS} edges. Two of them form the relevant conflict for the query: M1 states that Nobuhiro Watsuki is famous for \textit{The Fairly OddParents}, while M2 states that he is famous for \textit{Rurouni Kenshin}. Both memories assert the same predicate, ``is famous for,'' but provide different objects. M3 is also connected to the same entity, but it describes a different relation and should not compete as an answer candidate for this query. When the query was executed against this state, retrieval returned both M1 and M2. Because the retrieved context did not expose an explicit consolidation decision, the answer model selected \textit{Rurouni Kenshin}, which matches real-world knowledge but is incorrect under the benchmark's temporal ordering.

\paragraph{Dreaming consolidation.}

The middle panel of Figure~\ref{fig:dreaming-case-overview} summarizes the offline dreaming process. Dreaming starts from recently written or unconsolidated memories and expands over \texttt{MENTIONS} edges to form an entity-centered scope. In this case, the scope contains the conflicting pair M1--M2 together with the non-conflicting same-entity memory M3.

The first LLM pass performs relation detection and recognizes that M1 and M2 are competing answers to the same question: what Nobuhiro Watsuki is famous for. M3 is left outside the issue group because it describes a different relation rather than an alternative answer to this query. The second LLM pass performs action planning. Since M1 was ingested later than M2, the planner applies the benchmark's temporal supersede rule and selects M1 as the authoritative active memory. The resulting consolidation plan sets M2's status to \texttt{archived}, creates a \texttt{RELATED\_TO [supersedes]} edge from M1 to M2, and preserves consolidation provenance by linking the action to the source add records. The right panel of Figure~\ref{fig:dreaming-case-overview} shows the consolidated state: M1 remains active, M2 is archived, and the dashed \texttt{supersedes} edge records that the newer memory replaces the earlier conflicting memory.

\paragraph{Retrieval outcome.}

After dreaming, the same query was re-executed. Because archived memories are excluded from active retrieval, the answer model received only one active candidate for the ``famous for'' predicate: M1, which states that Nobuhiro Watsuki is famous for \textit{The Fairly OddParents}. Other retrieved memories, such as M3, may still be present through the shared entity, but they describe different relations and do not directly answer the query. With the conflicting earlier fact removed from the active context, the answer model produced \textit{The Fairly OddParents}, matching the benchmark ground truth. This case illustrates the main benefit of dreaming: the system does not rely on the answer model to compare timestamps or resolve contradictions during generation. Instead, temporal conflict resolution is converted into persistent memory state---an archived node plus a typed \texttt{supersedes} edge---so ordinary retrieval returns a cleaner context for this query.

\subsection{Feedback}

We examine a PersonaMem~\cite{jiang2025know} feedback-consistency item whose query asks: \textit{She has a free Saturday afternoon and wants a little social time without draining herself. Which invitation would she be most likely to say yes to?} Option A is an impromptu one-on-one thrift-store browse and coffee with no fixed end time and an easy exit if she gets tired. Option B is a relaxed board-game afternoon with two or three friends. The ground truth is Option A. Using memories generated by the MindVanilla memory-add pipeline, the answer model incorrectly selects Option B; after feedback, it selects the correct Option A.

\paragraph{Conversation evidence and feedback.}
The history shows that the user still wants social connection but prefers low-pressure, adjustable, and less group-intensive activities. For example, she says, ``If she asked me to just get tea one-on-one, I'd probably feel relieved. The big dinner is what feels impossible.'' She also notes that answering ``maybe'' and deciding at the last minute increases her anxiety and may appear flaky. The durable preference is therefore flexibility with a clear, manageable plan, rather than indiscriminately delaying decisions:
\begin{quote}\small
\textit{The user prefers low-pressure, adjustable social activities; compared with group gatherings, one-on-one plans with an easy exit are more manageable when her energy drops.}
\end{quote}

The user's concern about appearing unreliable or uncaring applies specifically when explaining or modifying plans with friends. Feedback therefore generates a conditioned communication memory without making the underlying social preference scenario-specific:
\begin{quote}\small
\textit{When helping the user explain social-planning preferences to friends, frame the need as a preference for flexible, low-pressure activities, not as unreliability or lack of care.}
\end{quote}

\begin{figure}[H]
    \centering
    \includegraphics[width=\linewidth,height=0.28\textheight,keepaspectratio]{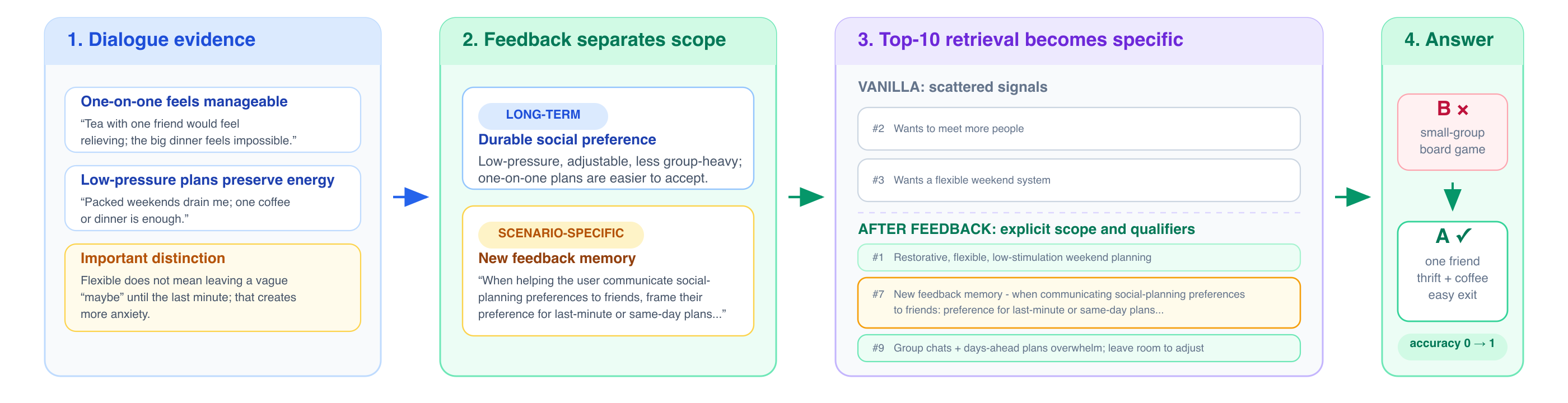}
    \caption{Feedback enriches the top-10 retrieval context and shifts the answer from Option B to the ground-truth Option A.}
    \label{fig:feedback-case-overview}
\end{figure}

\paragraph{Changes in top-10 retrieval.}
In the MindVanilla top-10, the most relevant memories state:
\begin{quote}\small
\textit{The user has been wanting to meet more people and get out of the house more.}\\
\textit{The user wants a simple weekend planning system that still feels flexible.}
\end{quote}
These memories capture social interest and flexibility separately, but do not clearly specify the preferred social format. The model consequently selects Option B, a relaxed but still group-based activity.

After feedback, the top-10 contains the new communication memory together with more specific evidence, including:
\begin{quote}\small
\textit{The user feels overwhelmed by constant group chats and social plans locked in days ahead, and would rather leave room to adjust.}\\
\textit{When helping the user decline or modify social plans, be warm and clear rather than leaving the decision as a vague ``maybe.''}
\end{quote}
Together, these memories provide a clearer decision boundary: the user wants social connection, but prefers one-on-one, low-pressure, adjustable, and easy-to-exit activities. Option A satisfies these conditions, whereas Option B remains group-based. The answer therefore changes from B to A, moving question-level correctness from 0 to 1. This single case illustrates the retrieval mechanism but does not by itself establish an aggregate benchmark gain.

\subsection{MindMemEvolve}
\label{sec:memevolve-case}

We evaluate MindMemEvolve on the PersonaMem benchmark introduced in Section~4.1. For this experiment, we first cluster the dialogue sessions by semantic similarity and randomly select 12 clusters, comprising 196 QA pairs, as the training set. The remaining 393 QA pairs constitute the test set. To isolate the effect of MindSchema and MindMemEvolve, we disable the episodic fallback mechanism and the default catch-all property, restricting memory extraction to schema-defined entities and properties. This ablation necessarily yields a lower pass rate than the full MindMemOS configuration.

\paragraph{Bootstrap schema.}
The bootstrap schema $\mathcal{S}_0$ is a minimal manually-authored template: a single \texttt{user} entity with one static property (\texttt{name}) and two dynamic properties---\texttt{mood\_feature} (inferred emotional state) and \texttt{habit\_feature} (inferred behavioral pattern). Despite its simplicity, this design is sufficiently expressive to seed the evolutionary process.

\paragraph{Evolution setup and results.}
MindMemEvolve runs for 5 epochs with 12 steps per epoch (one per cluster), using a population size of $K = 10$. On the training set, the bootstrap schema achieves $60.20\%$ accuracy; the best-evolved schema after epoch 5 reaches $64.28\%$.

\begin{figure}[H]
    \centering
    \includegraphics[width=0.48\linewidth]{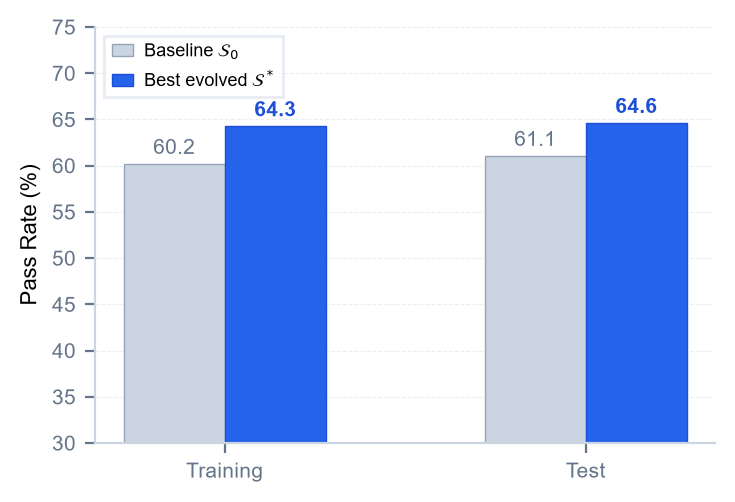}
    \caption{Training and test pass rates of MindMemEvolve on PersonaMem.}
    \label{fig:memevolve-bar}
\end{figure}

Best schema selection is based solely on training-set fitness. On the held-out test set of 393 QA pairs, the bootstrap schema achieves $61.07\%$, while the best-evolved schema attains $64.63\%$---a gain of $3.56$ percentage points. These results demonstrate modest but consistent improvement together with reasonable generalization. The limited margin is partly attributable to PersonaMem's reliance on inferential reasoning rather than pure historical information analysis; We are actively seeking more sensitivity-demanding scenarios to better elucidate the method's potential. Figure~\ref{fig:memevolve-bar} summarizes these findings.

\paragraph{Key evolved properties.}
The evolved schema expands its dynamic properties from 2 to 49, each targeting a narrow facet of user modeling. Two first-order properties and two higher-order properties are listed below to illustrate the mechanism.

\textbf{First-order properties} capture concrete episodes and explicit facts:
\texttt{activity\_participation\_record} documents the user's engagement in specific routine activities with full contextual detail (activity name, date, location, companions, role, outcome), and \texttt{creative\_cultural\_activity\_decision\_record} tracks explicit decisions to start, join, pause, or quit creative or cultural pursuits. These replace the monolithic \texttt{habit\_feature} of $\mathcal{S}_0$, whose unstructured text field conflated participation facts with activity preferences and creative decisions.

\textbf{Higher-order properties} synthesize recurring patterns across multiple first-order observations. \texttt{social\_energy\_management\_style} captures the user's enduring pattern for regulating social exposure, solitude, and recovery time---a trait that no single episode can reveal but that strongly informs personalized recommendations. \texttt{novelty\_seeking\_tendency} aggregates evidence across domains (learning, leisure, travel, food) into a stable cross-situational tendency. This first-order \textit{versus} higher-order distinction emerges autonomously during evolution: properties requiring multi-episode synthesis are automatically assigned $\text{order}=2$ through mutation operations, without manual annotation.

The full baseline and evolved schemas are provided in Appendix~\ref{app:memevolve-schema}.

\subsection{Skill Evolution}
\label{skill-evolution-case}

We compare the skill optimization results produced by \systemname{} on SpreadsheetBench. This case study focuses on two contrasts: the difference between the original skill and self-evolved skills, and the difference between unsupervised and supervised self-evolution. In the prompts, \textit{red} marks rules introduced by self-evolution relative to the original skill; in the supervised prompt, \textit{blue} further marks rules that appear only after supervised evolution, beyond the unsupervised version.

\paragraph{Original skill.}
The original skill only provides basic \texttt{openpyxl} and \texttt{pandas} usage and a small number of generic cautions. It tells the agent how to open, edit, and save a workbook, but it does not cover the failure modes that appear most often in SpreadsheetBench: formulas are not recalculated by \texttt{openpyxl}, \texttt{ws.max\_row} can be inflated by formatted empty rows, reading and writing on a live worksheet can move rows or overwrite source data, and the saved workbook must be reopened and verified in the same mode used by evaluation.

\paragraph{Unsupervised self-evolved skill.}
Unsupervised evolution uses execution traces but not task scores. Its changes mainly come from repeated workflow failures: the agent writes formulas and assumes that \texttt{wb.save()} has produced computed results; it treats formula text as numeric values; it reads and writes on a live worksheet during filtering, lookup, sorting, deletion, or expansion tasks; and it only checks whether the script ran successfully instead of reopening the output file for verification. Compared with the original skill, the unsupervised version distills these experiences into more concrete operational constraints, highlighted in \textit{red}.

\paragraph{Supervised self-evolved skill.}
Supervised evolution additionally uses task scores, so it not only summarizes recurring trace patterns but also separates rules that prevent task failures from rules that may damage user intent. Compared with the unsupervised version, the supervised version emphasizes two boundaries. The first is the boundary between formulas and hardcoded values, which prevents the skill from sacrificing explicitly requested reusable formulas merely to pass cached-value evaluation. The second is the semantic confirmation boundary before writing, which requires multiple concrete examples to confirm source columns, target columns, grouping structure, or mapping relations in complex spreadsheets. In this prompt, \textit{red} denotes evolved rules shared with the unsupervised prompt, while \textit{blue} denotes supervised-only additions.

\paragraph{Case analysis.}
From the original skill to unsupervised self-evolution, the main change is a shift from tool API instructions to executable error-avoidance rules. The \textit{red} rules directly correspond to failures that recur in spreadsheet tasks: empty formula caches, formatted rows that inflate \texttt{max\_row}, row movement on live worksheets that causes skipped or duplicated writes, merged cells that cannot be assigned directly, and missing post-save verification. In other words, unsupervised signals are already sufficient to summarize local failures in trajectories into general operational constraints.

From unsupervised to supervised evolution, the change is not simply that more rules are added; rather, the rules become more bounded. After observing task scores, the supervised version avoids overgeneralizing ``literal values are easier to pass cached-value checks'' into ``all formulas should be hardcoded.'' It therefore adds \textit{blue} supervised-only rules such as \textit{preserve reusable formulas}, \textit{distinguish formula-text grading from saved-value grading}, and \textit{state materialization limitations}. At the same time, it strengthens the pre-write check from a generic ``inspect workbook'' instruction into \textit{confirm mappings with at least two concrete source-to-target examples}, and it expresses dangerous operations as \textit{Blacklisted behavior}. This suggests that supervised signals mainly help the skill learn when to apply a rule, when to stop, and how to avoid sacrificing user intent for local scoring gains.

The full skill contents for all three versions are provided in Appendix~\ref{app:skills}.

\section{Conclusion and Future Work}

\subsection{Conclusion}

Memory is an important infrastructure component for persistent context management, personalization, and agent adaptation. In this work, we presented MindMemOS, a portable and self-evolving memory operating layer that integrates scenario-adaptive memory modeling, memory generation and retrieval, offline consolidation, and feedback-driven correction within a shared memory lifecycle. MindMemEvolve adapts memory schemas using task-specific validation signals, while MindSkillEvolve transforms accumulated execution trajectories into progressively refined and versioned skills. Through scenario-adaptive modeling and compact agentic search, MindMemOS achieves state-of-the-art performance under the reported evaluation protocols, particularly on multi-hop questions that require synthesizing evidence distributed across multiple memories. Dreaming and feedback further support the consolidation, correction, and traceable maintenance of accumulated information. Evaluations across dialogue memory, personalization, memory consolidation, and task execution, together with representative case studies, demonstrate the effectiveness of MindMemOS under the evaluated settings.

\subsection{Future Work}

In future work, we will focus on the following directions:




\paragraph{Experience-to-Skill Evolution.}
Future work will further connect experiential memory with the skill system by synthesizing execution trajectories, tool-use records, failures, and user corrections into reusable skills. These skills can be continuously validated and refined based on execution outcomes, improving task success and knowledge transfer across related tasks.

\paragraph{File-System-Level Memory Management.}
We also plan to incorporate files and documents as first-class memory artifacts, linking their content, metadata, versions, and provenance to the memory structure. This extension will support incremental indexing, cross-file reasoning, and traceable updates between source files and derived memories.
\section{Author Contributions}

\paragraph{Author List.}
Kaichao Liang\textsuperscript{1},
Yuqi Cui\textsuperscript{1},
Hao Kong\textsuperscript{1},
Xinyuan Huang\textsuperscript{1},
Guohaotian Hou\textsuperscript{1},
Qingcan Kang\textsuperscript{1},
Liang Chen\textsuperscript{1},
Yiyang Yin\textsuperscript{1},
Ke Ye\textsuperscript{1},
Jiaquan Guo\textsuperscript{1},
Da Chen\textsuperscript{1},
Xinduo Liu\textsuperscript{1},
Lingan Zeng\textsuperscript{1},
Yixing Peng\textsuperscript{1},
Rong Yao\textsuperscript{1},
Shixiong Kai\textsuperscript{1,*},
Mingxuan Yuan\textsuperscript{1,*}.

{\small \textsuperscript{1}Noah's Ark Lab, Huawei Technologies. \textsuperscript{*}Project Leader. Corresponding author}

\paragraph{Division of Work.}

\begin{itemize}
    \item \textbf{MindSchema:} Kaichao Liang, Yiyang Yin.
    \item \textbf{Scenario-Adaptive Memory Modeling:} Kaichao Liang.
    \item \textbf{Prototype:} Kaichao Liang, Yiyang Yin, Qingcan Kang, Lingan Zeng, Yixing Peng.
    \item \textbf{MindVanilla:} Hao Kong, Liang Chen.
    \item \textbf{Dreaming \& Feedback:} Yuqi Cui, Xinyuan Huang, Guohaotian Hou.
    \item \textbf{MindSkillEvolve:} Yuqi Cui, Ke Ye.
    \item \textbf{Product Design \& Engineering:} Yuqi Cui, Kaichao Liang, Hao Kong, Xinyuan Huang, Guohaotian Hou, Qingcan Kang, Liang Chen, Yiyang Yin, Ke Ye, Jiaquan Guo, Da Chen, Xinduo Liu, Lingan Zeng, Yixing Peng, Rong Yao.
    \item \textbf{Technical Leads:} Shixiong Kai, Mingxuan Yuan.
\end{itemize}

\bibliographystyle{plain}
\newpage
\bibliography{references}

\newpage
\appendix
\section{Schema Definitions for MindMemEvolve Case Study}
\label{app:memevolve-schema}

\subsection*{Baseline Schema (\texorpdfstring{$\mathcal{S}_0$}{S0})}

\begin{promptbox}{Baseline Schema \texorpdfstring{$\mathcal{S}_0$}{S0} (2 dynamic properties)}{prompt:schema-baseline}
{\ttfamily\footnotesize
\{\par
\ \ "entity\_type": "user",\par
\ \ "entity\_instruction": "Two kinds of properties: XXX\_event / XXX\_record = [Factual Record]; XXX\_feature = [Analytical Summary]. Infer implicit attitudes from behavioral signals.",\par
\ \ "static\_property": \{ "name": "User name or identifier" \},\par
\ \ "dynamic\_property": \{\par
\ \ \ \ "first\_order": [\par
\ \ \ \ \ \ "mood\_feature",\par
\ \ \ \ \ \ "habit\_feature"\par
\ \ \ \ ]\par
\ \ \}\par
\}\par
}
\end{promptbox}

\subsection*{Evolved Schema (\texorpdfstring{$\mathcal{S}^*$}{S*})}

\begin{promptbox}{Evolved Schema \texorpdfstring{$\mathcal{S}^*$}{S*} --- 49 dynamic properties}{prompt:schema-evolved}
{\ttfamily\footnotesize
\{\par
\ \ "entity\_type": "user",\par
\ \ "entity\_instruction": "Maintain exactly one user entity per conversation. Properties are divided into first\_order (single-episode facts) and higher\_order (cross-episode syntheses). Every property value must be an atomic, evidence-grounded, semantically complete sentence.",\par
\ \ "static\_property": \{ "name": "User name or identifier" \},\par
\ \ "dynamic\_property": \{\par
\ \ \ \ // --- First-order: single-episode facts (41 properties) ---\par
\ \ \ \ "first\_order": [\par
\ \ \ \ \ \ "activity\_participation\_record",\par
\ \ \ \ \ \ "creative\_cultural\_activity\_decision\_record",\par
\ \ \ \ \ \ "participation\_format\_preference\_feature",\par
\ \ \ \ \ \ "activity\_fit\_preference\_feature",\par
\ \ \ \ \ \ "explicit\_preference\_transition\_record",\par
\ \ \ \ \ \ "shared\_activity\_transition\_record",\par
\ \ \ \ \ \ "leisure\_environment\_fit\_feature",\par
\ \ \ \ \ \ "recommendation\_relevance\_profile\_feature",\par
\ \ \ \ \ \ "engagement\_reward\_driver\_feature",\par
\ \ \ \ \ \ "domain\_social\_exchange\_orientation\_feature",\par
\ \ \ \ \ \ "pursuit\_engagement\_record",\par
\ \ \ \ \ \ "interest\_engagement\_trajectory\_record",\par
\ \ \ \ \ \ "narrative\_interpretive\_engagement\_feature",\par
\ \ \ \ \ \ "pursuit\_engagement\_lifecycle\_record",\par
\ \ \ \ \ \ "creative\_format\_transition\_record",\par
\ \ \ \ \ \ "participation\_role\_comfort\_feature",\par
\ \ \ \ \ \ "pursuit\_lifecycle\_record",\par
\ \ \ \ \ \ "supportive\_discussion\_norm\_preference\_feature",\par
\ \ \ \ \ \ "learning\_engagement\_preference\_feature",\par
\ \ \ \ \ \ "expressive\_processing\_fit\_feature",\par
\ \ \ \ \ \ "creative\_life\_feature",\par
\ \ \ \ \ \ "learning\_format\_affinity\_feature",\par
\ \ \ \ \ \ "peer\_learning\_collaboration\_record",\par
\ \ \ \ \ \ "creative\_structure\_spontaneity\_fit\_feature",\par
\ \ \ \ \ \ "creative\_workflow\_preference\_feature",\par
\ \ \ \ \ \ "learning\_preference\_feature",\par
\ \ \ \ \ \ "narrative\_media\_self\_reflection\_feature",\par
\ \ \ \ \ \ "learning\_method\_trial\_record",\par
\ \ \ \ \ \ "contextual\_interpretive\_analysis\_feature",\par
\ \ \ \ \ \ "feedback\_motivation\_sensitivity\_feature",\par
\ \ \ \ \ \ "creative\_project\_record",\par
\ \ \ \ \ \ "expressive\_feedback\_impact\_record",\par
\ \ \ \ \ \ "creative\_critique\_orientation\_feature",\par
\ \ \ \ \ \ "knowledge\_sharing\_initiative\_record",\par
\ \ \ \ \ \ "informational\_learning\_impact\_record",\par
\ \ \ \ \ \ "information\_response\_record",\par
\ \ \ \ \ \ "information\_effect\_record",\par
\ \ \ \ \ \ "personal\_health\_status\_record",\par
\ \ \ \ \ \ "child\_development\_guidance\_motivation\_feature",\par
\ \ \ \ \ \ "residence\_situation\_record",\par
\ \ \ \ \ \ "work\_education\_status\_record"\par
\ \ \ \ ],\par
\par
\ \ \ \ // --- Higher-order: cross-episode syntheses (8 properties) ---\par
\ \ \ \ "higher\_order": [\par
\ \ \ \ \ \ "social\_energy\_management\_style",\par
\ \ \ \ \ \ "novelty\_seeking\_tendency",\par
\ \ \ \ \ \ "commitment\_sustainability\_pattern",\par
\ \ \ \ \ \ "self\_reflection\_orientation",\par
\ \ \ \ \ \ "decision\_making\_style",\par
\ \ \ \ \ \ "resilience\_adaptation\_pattern",\par
\ \ \ \ \ \ "autonomy\_orientation",\par
\ \ \ \ \ \ "generativity\_orientation"\par
\ \ \ \ ]\par
\ \ \}\par
\}\par
}
\end{promptbox}
\newpage
\section{Full Skill Contents for Skill Evolution Case Study}
\label{app:skills}

This appendix provides the complete skill contents referenced in the Skill Evolution case study (Section~\ref{skill-evolution-case}). Text in \skillred{red} marks rules introduced by self-evolution relative to the original skill; in the supervised prompt, \skillblue{blue} further marks rules that appear only after supervised evolution.

\begin{promptbox}{Initial Skill for SpreadsheetBench}{prompt:skill-original-xlsx}
\textit{---}\\
\textit{name: xlsx}\\
\textit{description: Use this skill whenever the user wants to do anything with Excel spreadsheet files (.xlsx, .xls, .csv). This includes reading data, writing formulas, manipulating cells, formatting, filtering, creating charts, pivot tables, and any spreadsheet automation tasks.}\\
\textit{---}\\
\# Excel Spreadsheet Processing\\
\par
Use `openpyxl` to read and write .xlsx files.\\
\par
\#\# Quick Start\\
```python\\
from openpyxl import load\_workbook\\
\par
wb = load\_workbook("input.xlsx")\\
ws = wb["Sheet1"]\\
\par
value = ws["A1"].value\\
ws["B2"] = 42\\
ws["C2"] = "=SUM(A2:B2)"\\
\par
wb.save("output.xlsx")\\
```\\
\par
Use this for direct cell edits, formula updates, and simple workbook changes.\\
\par
\#\# Reading Data with pandas\\
```python\\
import pandas as pd\\
\par
df = pd.read\_excel('file.xlsx')                          \# First sheet\\
all\_sheets = pd.read\_excel('file.xlsx', sheet\_name=None) \# All sheets as dict\\
```\\
\par
\#\# Common Pitfalls\\
- **Cell indices are 1-based**: `ws.cell(row=1, column=1)` is A1.\\
- **`data\_only=True` destroys formulas on save**: Use a separate workbook object for reading calculated values.\\
- **`ws.max\_row` overcounts**: May include formatted-but-empty rows. Scan the column to find the last non-empty cell when you need the true data range.
\end{promptbox}

\begin{promptbox}{Evolved Skill by MindMemOS-unsup}{prompt:skill-unsupervised-xlsx}
\textit{---}\\
\textit{name: xlsx}\\
\textit{description: Use this skill whenever the user wants to do anything with Excel spreadsheet files (.xlsx, .xls, .csv). This includes reading data, writing formulas, manipulating cells, formatting, filtering, creating charts, pivot tables, and any spreadsheet automation tasks.}\\
\textit{---}\\
\# Excel Spreadsheet Processing\\
\par
Use `openpyxl` to read and write .xlsx files.\\
\skillred{If the workbook will be judged from saved/cached values, write computed literal values in Python. Only write formulas when you know Excel will recalculate them on open. Never assume wb.save() computes formulas.}\\
\skillred{High-risk action: do not silently replace a user-requested reusable formula with hardcoded values unless the task is explicitly graded from saved cached values or recalculation is unavailable.}\\
\skillred{- **High-risk action: do not leave formula cells as the only source of truth when the workbook will be checked from saved/displayed values.** If you cannot guarantee recalculation on open, compute the result in Python and write literal values instead.}\\
\skillred{If the task will be judged from the saved workbook's displayed/cached values, write the final computed value directly into the cell. After saving, reopen with `data\_only=True` to confirm the stored value is present.}\\
\par
\#\# Quick Start\\
```python\\
from openpyxl import load\_workbook\\
wb = load\_workbook("input.xlsx")\\
ws = wb["Sheet1"]\\
value = ws["A1"].value\\
ws["B2"] = 42\\
ws["C2"] = "=SUM(A2:B2)"\\
wb.save("output.xlsx")\\
```\\
\par
Use this for direct cell edits, formula updates, and simple workbook changes.\\
\skillred{High-risk action: do not mutate openpyxl style objects in place; create a new `PatternFill`/`Border` or `copy()` the style, then assign it back to the cell.}\\
\skillred{For filtering, lookup, mapping, row moves, deduping, or sorting, first read the full source table into Python, confirm headers and the true data extent, compute the result in memory, and write once.}\\
\skillred{If source cells may contain formulas, read them in two separate passes: one normal load for formula text, and one `data\_only=True` load for cached results. Never use the same `.value` as both a formula string and a computed input.}\\
\skillred{Do not replace a requested dynamic spreadsheet behavior with one-off hardcoded outputs; this can make current cells look correct but break the workbook when inputs change.}\\
\skillred{- **For deletions, insertions, row duplication, or "expand each row" tasks, read the full source rows into Python, compute the transformed table in memory, then clear and rewrite the destination range once.**}\\
\par
\#\# Before Editing\\
\skillred{- **Before writing into a target range, inspect `ws.merged\_cells.ranges`.** Never assign to a merged placeholder cell (`MergedCell`).}\\
\skillred{- Do not rely on `ws.max\_row` when empty formatted rows may exist. Scan upward from the bottom to find the last row with any non-empty relevant cell.}\\
\skillred{- If the target sheet, range, or transformation pattern is not fully explicit, do a read-only inspection pass first and confirm the real data extent, headers, and any example mappings before writing.}\\
\skillred{- After saving, reopen the file twice when needed: once normally to confirm formulas/styles, and once with `data\_only=True` to confirm displayed values.}\\
\par
\#\# Common Pitfalls\\
- **Cell indices are 1-based**: `ws.cell(row=1, column=1)` is A1.\\
- **`data\_only=True` destroys formulas on save**: Use a separate workbook object for reading calculated values.\\
\skillred{- **Never treat a formula cell's `.value` as a plain literal when the workbook contains formulas.**}\\
\skillred{- Do not trust `ws.max\_row` alone. To find the real data extent, scan upward from the bottom until the last row with any non-empty cell.}\\
\skillred{- If the task includes formatting or structural edits, verify those changes with a normal workbook reload; `data\_only=True` only confirms values, not formatting.}\\
\skillred{- Do not use a `data\_only=True` workbook as the source for edits.}
\end{promptbox}

\begin{promptbox}{Evolved Skill by MindMemOS-sup}{prompt:skill-supervised-xlsx}
\textit{---}\\
\textit{name: xlsx}\\
\textit{description: Use this skill whenever the user wants to do anything with Excel spreadsheet files (.xlsx, .xls, .csv). This includes reading data, writing formulas, manipulating cells, formatting, filtering, creating charts, pivot tables, and any spreadsheet automation tasks.}\\
\textit{---}\\
\# Excel Spreadsheet Processing\\
\par
Use `openpyxl` to read and write .xlsx files.\\
\skillred{If the workbook will be judged from saved/cached values, write computed literal values in Python. Only write formulas when you know Excel will recalculate them on open. Never assume wb.save() computes formulas.}\\
\skillred{High-risk action: do not silently replace a user-requested reusable formula with hardcoded values unless the task is explicitly graded from saved cached values or recalculation is unavailable.}\\
\skillblue{**If the user explicitly asks for a reusable formula, do not silently replace it with hardcoded values.**}\\
\skillblue{Before writing, decide whether the grader will check formula text or saved cell values. If cached values are graded and recalculation on open is not confirmed, materialize literal results in Python and state that limitation in the final note. If reusable formulas are explicitly required, preserve formulas.}\\
\skillblue{If reusable formulas are required and recalculation on open is not confirmed, do not silently materialize values; instead, write the formula, verify it separately with a `data\_only=True` read, and state that cached results may not update until Excel recalculates.}\\
\skillblue{**If the user asks to fix or rewrite formulas, do not silently replace them with literal values just because cached values are graded.**}\\
\skillblue{If the user's instruction or example could be interpreted more than one way, do a read-only check against at least 2 concrete examples and resolve the exact semantics before writing.}\\
\par
\#\# Quick Start\\
```python\\
from openpyxl import load\_workbook\\
wb = load\_workbook("input.xlsx")\\
ws = wb["Sheet1"]\\
value = ws["A1"].value\\
ws["B2"] = 42\\
ws["C2"] = "=SUM(A2:B2)"\\
wb.save("output.xlsx")\\
```\\
\par
Use this for direct cell edits, formula updates, and simple workbook changes.\\
\skillblue{For any nontrivial write, do a read-only pass that confirms: (a) the true used range, (b) the header row, (c) merged cells, and (d) at least 2 concrete source-to-target examples.}\\
\skillblue{For any transformation that changes row order, filters rows, dedupes, expands rows, or maps one table to another, first do a read-only inspection pass to confirm the true data bounds, header row, merged cells, and at least 2 concrete source-to-target examples.}\\
\skillred{High-risk action: do not mutate openpyxl style objects in place; create a new `PatternFill`/`Border` or `copy()` the style, then assign it back to the cell.}\\
\skillred{For filtering, lookup, mapping, row moves, deduping, sorting, or flattening tasks, first do a read-only inspection pass, then read the full source table into Python, compute the final output in memory, and write once.}\\
\skillblue{**Blacklisted behavior:** iterating over a live worksheet while simultaneously reading and writing rows/cells for filtering, lookup, deduping, sorting, row moves, deletion/insertion, or row expansion.}\\
\skillblue{**Blacklisted behavior:** replacing a user-requested reusable formula with hardcoded values, or writing a formula when the task is explicitly graded as final values only, without stating the limitation.}\\
\skillred{**If a source column may contain formulas, inspect it in two separate passes: load once normally to see formula text, and once with `data\_only=True` to see cached results.}\\
\skillblue{Before looking up or matching values, verify that the chosen key column actually contains the intended identifiers in the workbook, not a similarly named but different field.}\\
\skillblue{High-risk action: do not generate or extend formulas by naive string replacement on existing formula text; build each target formula explicitly from the row number.}\\
\skillblue{After saving, reopen the workbook in the mode the grader will use and verify at least 2--3 representative source-to-target pairs by actual value or formula text.}\\
\par
\#\# Before Editing\\
\skillblue{**For any task that depends on row/column relationships, month/category grouping, block boundaries, or lookup criteria, do a read-only inspection pass first and confirm the real headers, data extent, and at least 2 concrete source-to-target examples before writing.**}\\
\skillblue{**If the target sheet, range, header row, or data extent is not fully explicit, do a read-only inspection pass first and confirm the real headers, used range, and any merged cells before writing.**}\\
\skillblue{High-risk action: do not infer the transformation from a single sample cell, nearby labels, or visible formatting alone.}\\
\skillblue{**Before any transformation, inspect the actual source rows/columns and confirm the real header row, data extent, and at least 2 concrete source-to-target examples.**}\\
\skillred{**Before writing, inspect `ws.merged\_cells.ranges`. Do not assign into a merged placeholder cell (`MergedCell`).**}\\
- If a named sheet, cell, column, header, row, or target range is missing or ambiguous, stop and inspect the workbook contents before editing.\\
- If the target sheet, range, or transformation pattern is not fully explicit, do a read-only inspection pass first and confirm the real data extent, headers, and any example mappings before writing.\\
- After reopening the output, verify that the result matches the intended transformation on at least 2--3 representative examples from the source.\\
\par
\#\# Common Pitfalls\\
- **Cell indices are 1-based**: `ws.cell(row=1, column=1)` is A1.\\
- **`data\_only=True` destroys formulas on save**: Use a separate workbook object for reading calculated values.\\
\skillred{- **Never treat a formula cell's `.value` as a plain literal when the workbook contains formulas.**}\\
\skillred{- Do not trust `ws.max\_row` alone. To find the real data extent, scan upward from the bottom until the last row with any non-empty cell.}
\end{promptbox}
\newpage
\section{MindMemEvolve Algorithms}
\label{app:algorithms}

\begin{algorithm}[H]
\caption{MindMemEvolve --- LLM-Guided Evolutionary Schema Optimization}
\label{alg:mindmemevolve}
\begin{algorithmic}[1]

\Require Training set $\mathcal{D} = \{(\mathbf{context}_i, q_i, a_i)\}_{i=1}^{N}$, initial schema $\mathcal{S}_0$, epochs $E$, clusters $U$, population size $K$, elite fraction $\alpha$, prune prob.\ $p_{\text{prune}}$, crossover prob.\ $p_{\text{cross}}$
\Ensure Approximately optimal schema $\mathcal{S}^*$

\State $H \gets E \cdot U$ \Comment{Total cumulative training steps}
\State $\mathbb{C} \gets \Call{PartitionClusters}{\mathcal{D}, U}$ \Comment{Partition $\mathcal{D}$ into $U$ clusters}

\For{$j \gets 1$ \textbf{to} $K$}
    \State $\mathcal{S}_{1,j} \gets \Call{RandomMutate}{\mathcal{S}_0}$ \Comment{$K$ diversified individuals from $\mathcal{S}_0$}
\EndFor

\For{$e \gets 1$ \textbf{to} $E$}
    \For{$u \gets 1$ \textbf{to} $U$}
        \State $t \gets (e-1) \cdot U + u$ \Comment{Cumulative step index}

        \For{$j \gets 1$ \textbf{to} $K$}
            \State $\Call{InitSandbox}{\mathcal{S}_{t,j}}$ \Comment{Clean memory store}
            \State $\Call{AddMemories}{\mathbb{C}_u.\!\mathbf{contexts}, \mathcal{S}_{t,j}}$ \Comment{Ingest cluster $u$}
            \State $\mathbf{s} \gets []$
            \ForAll{$(q_i, a_i) \in \mathbb{C}_u$}
                \State $\mathbf{r}_i \gets \Call{SearchMemories}{q_i, \mathcal{S}_{t,j}}$
                \State $\hat{a}_i \gets \Call{GenerateAnswer}{q_i, \mathbf{r}_i}$
                \State $\Call{AppendScore}{\mathbf{s}, a_i, \hat{a}_i}$
            \EndFor
            \State $f(\mathcal{S}_{t,j}) \gets \Call{Mean}{\mathbf{s}}$
        \EndFor

        \State $\Call{SortDescending}{\{\mathcal{S}_{t,1}, \dots, \mathcal{S}_{t,K}\}, f}$
        \State $n_{\text{elite}} \gets \lceil \alpha \cdot K \rceil$

        \For{$j \gets 1$ \textbf{to} $n_{\text{elite}}$}
            \State $\mathcal{S}_{t+1,j} \gets \mathcal{S}_{t,j}$ \Comment{Elite preservation}
        \EndFor

        \For{$j \gets n_{\text{elite}} + 1$ \textbf{to} $K$}
            \State $(\mathcal{P}_a, \mathcal{P}_b) \gets \Call{TournamentSelect}{\{\mathcal{S}_{t,j}\}_{j=1}^{K}}$

            \ForAll{$\mathcal{P} \in \{\mathcal{P}_a, \mathcal{P}_b\}$}
                \State $\mathcal{P} \gets \Call{InducedMutate}{\mathcal{P}, \mathbb{C}_u}$
                \State $\mathcal{P} \gets \Call{InjectPlausibleTypes}{\mathcal{P}}$
                \State $\mathcal{P} \gets \Call{PruneLowFrequency}{\mathcal{P}, p_{\text{prune}}}$
            \EndFor

            \State $\mathcal{S}_{t+1,j} \gets \Call{CrossoverInsert}{\mathcal{P}_a, \mathcal{P}_b, p_{\text{cross}}}$
        \EndFor
    \EndFor
\EndFor

\For{$j \gets 1$ \textbf{to} $K$}
    \State $f(\mathcal{S}_{H+1,j}) \gets$ evaluate $\mathcal{S}_{H+1,j}$ on full training set $\mathcal{D}$ \Comment{Over all $N$ samples}
\EndFor
\State $\mathcal{S}^* \gets \arg\max_{j \in [1,K]} \; f(\mathcal{S}_{H+1,j})$
\State \Return $\mathcal{S}^*$

\end{algorithmic}
\end{algorithm}

\end{document}